\documentclass[review,10pt]{JMtemplate}
\usepackage{bm}
\usepackage{graphicx} 
\usepackage{amsmath,mathrsfs,amssymb} 
\usepackage{amsfonts}  % blackboard math symbols, e.g., \mathbb
\usepackage{algorithm}
\usepackage{algorithmic}
\usepackage{siunitx}   % 用于书写 数字+单位 的格式
\usepackage{upgreek}   % 用于书写 数字+单位 的格式
\usepackage{xcolor}    % colors
\usepackage{subfigure}
\usepackage{booktabs}      % professional-quality tables
\usepackage{longtable}
\usepackage{threeparttable}
\usepackage{multirow}
\usepackage{multicol}
\usepackage{times}
\usepackage{helvet}
\usepackage{courier}
\usepackage{marvosym}
\usepackage[hyphens]{url} 
\usepackage{natbib}  % DO NOT CHANGE THIS AND DO NOT ADD ANY OPTIONS TO IT
\usepackage{caption} % DO NOT CHANGE THIS AND DO NOT ADD ANY OPTIONS TO IT 
\usepackage{adjustbox}
\usepackage{subcaption}
\usepackage{subfigure}

\newcommand{\zhangsqchecked}[1] {\textcolor{black}{#1}}

\usepackage[
    colorlinks=true,    % 启用彩色链接
    linkcolor=blue,     % 内部链接颜色
    urlcolor=blue,      % URL链接颜色
    citecolor=green,    % 引用链接颜色
    filecolor=magenta   % 文件链接颜色
]{hyperref}
\usepackage{textcomp}

\begin{document}
\begin{frontmatter}
\title{TernaryCLIP: Efficiently Compressing Vision-Language Models with Ternary Weights and Distilled Knowledge}

\author{Shu-Hao Zhang\textsuperscript{1}, 
Wei-Cheng Tang\textsuperscript{1}, 
Chen Wu\textsuperscript{2}, 
Peng Hu\textsuperscript{2}, 
Nan Li\textsuperscript{2}, 
Liang-Jie Zhang\textsuperscript{2}, \\
Qi Zhang\textsuperscript{2}, 
Shao-Qun Zhang\textsuperscript{1, \Letter}}
\address{\textsuperscript{1} State Key Laboratory of Novel Software Technology, Nanjing University, Nanjing 210063, China \\
\textsuperscript{2} Microsoft AI, Beijing 100080, China}

% 摘要：Lightweight 重点是 TernaryCLIP 的 Effectiveness + Efficiency 
\begin{abstract}
Recent years have witnessed an increasing interest in image-text contrastive modeling, exemplified by models such as Contrastive Language-Image Pretraining (CLIP). In this paper, we propose the TernaryCLIP, a lightweight computational framework that converts connection weights of both vision and text encoders of CLIP into the ternary format, instead of full-precision or floating ones. TernaryCLIP incorporates quantization-aware training and distillation modules, preventing precision degradation and enabling low-cost and high-efficiency computations. Comprehensive experiments demonstrate that TernaryCLIP can achieve up to 99\% ternarized weights with 1.58-bit representation, 16.98 $\times$ compression ratio, 2.3 $\times$ inference acceleration, 16 $\times$ storage reduction, 10 $\times$ memory optimization, and 60\% sparsity while maintaining promising performance on zero-shot image classification and image-text retrieval tasks across 41 commonly used datasets. Our work highlights the feasibility of extreme quantization for large multimodal models, supporting effective and efficient deployment on resource-constrained devices. The model and code can be accessed from \href{https://huggingface.co/xxx/TernaryCLIP_ViT-B-16}{Hugging Face} and \href{https://github.com/zhangsq-nju/TernaryCLIP}{GitHub}.

\end{abstract}
\end{frontmatter}

\begin{figure*}[!htt]
    \centering
    \includegraphics[width=\textwidth]{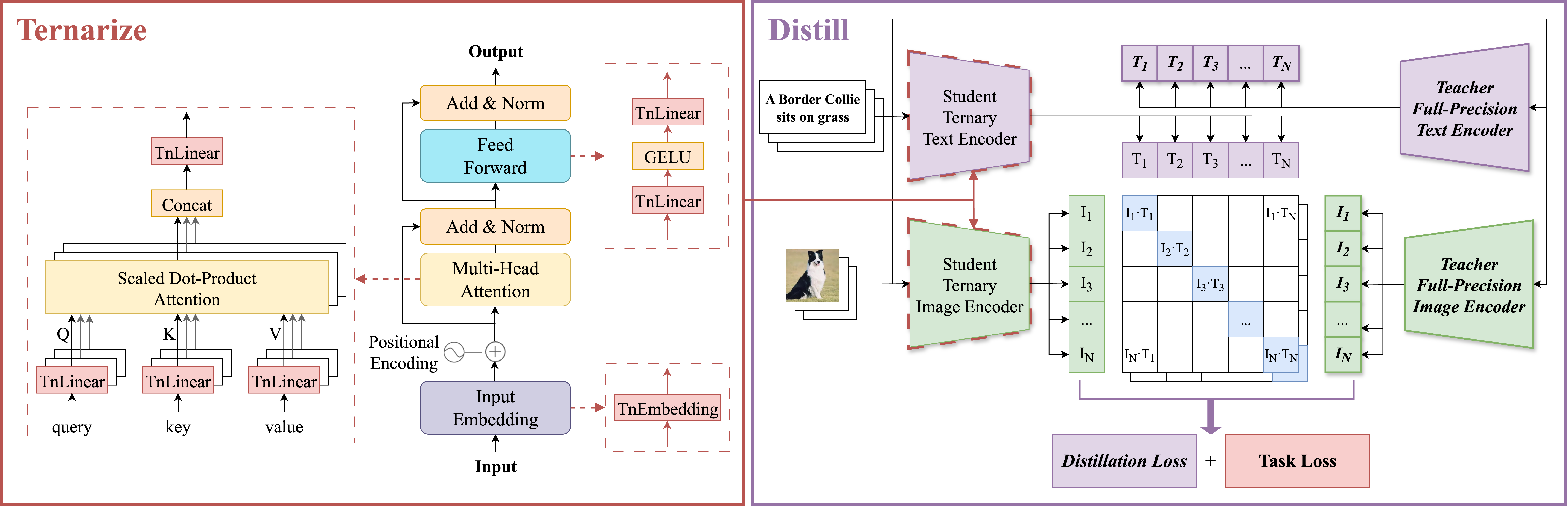}
    % \caption{Summary of TnCLIP: 1) Knowledge distillation of CLIP, 2) Ternarization-aware training of CLIP, 3) Distill a ternary CLIP.}
    \caption{The workflow of the TernaryCLIP framework, comprising two modules: ternarization and distillation.}
    \label{fig:tnclip}
    \vspace{-1.0em}
\end{figure*}

% Introduction -------------------------------------------------------
\section{Introduction}
Large-scale multimodal models, such as Contrastive Language-Image Pretraining (CLIP)~\cite{radford2021learning} that aligns images and texts in a shared embedding space through contrastive learning, have demonstrated exceptional performance across vision-language tasks, including image classification and image-text retrieval.
However, the impressive capabilities of these models come at the cost of significant resource demands~\cite{brown2020language, strubell2019energy}, contributing to three critical challenges for practical deployment.
First, the tremendous parameters result in substantial storage and memory consumption~\cite{han2016deepcompression}. Second, the full-precision model introduces heavy computational burdens affecting inference latency~\cite{sze2017efficientprocessingdeepneural}. Third, the deployment of downstream tasks typically requires massive, domain-specific, labeled datasets~\cite{devlin2019bert}, such as LAION-400M~\cite{schuhmann2021laion400m}, which incur huge annotation costs and limit practical applicability~\cite{changpinyo2021cc12m}.
These challenges hinder the widespread deployment of large-scale multimodal models in resource-constrained environments such as mobile devices~\cite{howard2017mobilenets}. Consequently, model lightweight techniques have emerged as a prominent research direction to bridge the gap between promising performance and feasible deployment~\cite{zhu2024surveymodelcompressionlarge}.

Quantization has emerged as a promising lightweight technique that converts full-precision weights into lower-bit formats~\cite{banner2019post, jacob2018quantization, wu2016quantized}. Specifically, ternary quantization compresses floating-point or 32-bit weights into three discrete values $\{-\Delta, 0, +\Delta\}$, reducing the storage and memory consumption via an ultra-low 1.58-bit format with each weight requiring $\log_2(3) \approx 1.585$ bits in the optimal case. This quantization enables the replacement of complex floating-point matrix multiplications with simple addition and shift operations, thereby improving inference efficiency and reducing latency~\cite{alemdar2017ternaryneuralnetworks, li2022ternaryweightnetworks, zhu2017trained}.
However, existing quantization approaches suffer from two critical limitations for multimodal efficient deployment. On the one hand, current research primarily focuses on unimodal architectures such as BERT for natural language processing~\cite{zafrir2019q8bert, fan2020ternarybert} and ViT for computer vision~\cite{li2022qvit, liu2021post}, leaving multimodal models like CLIP largely unexplored. On the other hand, quantization usually causes severe performance degradation that compromises practical applicability~\cite{banner2019post, rastegari2016xnor}.

Knowledge distillation has emerged as another lightweight technique that employs the teacher-student framework without massive labeled data~\cite{gou2021knowledge, hinton2015distilling}. Specifically, distillation resolves the annotation requirement challenge by utilizing soft labels from teacher models to guide student learning, transferring rich representational knowledge without relying on domain-specific labeled datasets~\cite{romero2015fitnets, zagoruyko2016paying}. This preserves the zero-shot capabilities of pre-trained models~\cite{radford2021learning}, thereby avoiding dependence on extensive labeled datasets and significantly reducing annotation costs. Moreover, distillation addresses training resource consumption by leveraging pre-trained teacher knowledge to substantially reduce student model training time, avoiding the enormous computational overhead of training from scratch~\cite{cho2019efficacy, mirzadeh2020improved}. Despite the promising progress, comprehensive distillation strategies for ultra-low-bit multimodal scenarios, such as ternary quantization, remain unexplored~\cite{mishra2017apprentice, polino2018model}.

In this paper, we propose the TernaryCLIP by integrating ternary quantization and distillation into large-scale vision-language models. The workflow of our TernaryCLIP is illustrated in Figure~\ref{fig:tnclip}. The experimental results indicate substantial improvements in reducing the resource demands. For computational efficiency, TernaryCLIP achieves 1.58-bit weight representation with 99\% quantization proportion and 16.98× compression ratio, resulting in 2.3× inference speedup, 16× storage reduction, 10× memory optimization, and 60\% sparsity. For annotation dependency elimination, TernaryCLIP maintains competitive zero-shot performance across 41 datasets, preserving the generalization capabilities of full-precision models without requiring extensive downstream task-specific annotations. These results demonstrate the effectiveness of our proposed TernaryCLIP, maintaining competitive performance with substantially reduced resource requirements.

To the best of our knowledge, the proposed TernaryCLIP is the first advance that achieves extremely-low compression of weight precision of large-scale vision-language models. Empirical results support the practicability that ultra-low-bit multimodal models are not only feasible but also capable of bridging the gap between performance and practical deployment in resource-constrained environments such as edge computing platforms.

% 1.58-bit\footnote{The 1.58-bit representation is the theoretical encoding requirement for ternary weights, where each weight can take one of three values $\{-1, 0, +1\}$, requiring $\log_2(3) \approx 1.585$ bits per weight in the optimal case.}

% --------------------------------------------------------------------

% Preliminaries -------------------------------------------------------
\section{Preliminaries}

\subsection{Introduction to CLIP}
The CLIP architecture~\citep{radford2021learning} comprises an image encoder $f_i$ (e.g., ViT) and a text encoder $f_t$ (e.g., BERT), which are trained to maximize cosine similarities between matched image-text pairs while minimizing similarities for unmatched pairs. Given a training dataset $D={(I_i,T_i)}_{i=1}^{|D|}$, CLIP learns the cross-modal representations through an InfoNCE-based contrastive loss~\citep{oord2019rlcpc}. TernaryCLIP adopts this contrastive learning framework with the loss function formulated as $\mathcal{L}_{\text{task}} = (\mathcal{L}_{\text{I} \rightarrow \text{T}} + \mathcal{L}_{\text{T} \rightarrow \text{I}}) / 2$, where $\mathcal{L}_{\text{I} \rightarrow \text{T}} = \text{CrossEntropy}(\text{logits}, \text{labels})$ and $\mathcal{L}_{\text{T} \rightarrow \text{I}} = \text{CrossEntropy}(\text{logits}^T, \text{labels})$. The logits are computed as $\text{logits} = I_e T_e^{\top} / \tau = f_i(I) f_t(T)^{\top} / \tau$, representing the scaled cosine similarities between image and text embeddings with a temperature hyperparameter $\tau$.

Inspired by previous distillation methods for CLIP, such as CLIP-KD~\citep{yang2024clip} and ComKD-CLIP~\citep{chen2024comkdclip}, TernaryCLIP integrates three complementary distillation strategies: Contrastive Relational Distillation (CRD), Interactive Contrastive Learning (ICL), and Feature Distillation (FD). CRD transfers contrastive knowledge from teacher to student by aligning their cross-modal similarity distributions through KL divergence: $\mathcal{L}_{\text{crd}} = \text{KL}(p_t | p_s) + \text{KL}(q_t | q_s)$, where $p_s, p_t$ and $q_s, q_t$ denote the image-to-text and text-to-image distributions for the student and teacher models, respectively. ICL facilitates cross-modal knowledge transfer by establishing contrastive relationships between student and teacher embeddings across modalities: $\mathcal{L}_{\text{icl}} = (\mathcal{L}_{I_s \rightarrow T_t} + \mathcal{L}_{T_s \rightarrow I_t}) / 2$, where $\text{logits}_{I_s \rightarrow T_t} = I_{e,s} T_{e,t}^{\top} / \tau$ and $\text{logits}_{T_s \rightarrow I_t} = T_{e,s} I_{e,t}^{\top} / \tau$ represent the scaled similarities between student image embeddings and teacher text embeddings, and between student text embeddings and teacher image embeddings, respectively. FD aligns the embedding spaces by minimizing the mean squared error between student and teacher representations: $\mathcal{L}_{\text{fd}} = \text{MSE}(I_{e,s}, I_{e,t}) + \text{MSE}(T_{e,s}, T_{e,t})$, where $\text{MSE}(I_{e,s}, I_{e,t}) = \sum_{i=1}^{n}(I_{e,s}^{(i)} - I_{e,t}^{(i)})^2 / n$ measures the distance between student and teacher image embeddings, and similarly for text embeddings.
% --------------------------------------------------------------------

% Related Work -------------------------------------------------------
\subsection{Related Studies}

\textbf{Model Quantization.}
Quantization has emerged as a fundamental strategy for compressing neural networks to enable efficient inference. Traditional approaches, such as integer quantization, reduce memory usage and computational costs while maintaining acceptable accuracy~\cite{jacob2018quantization, wu2016quantized}. Recent advances have extended to ternary quantization, which represents weights using only three values $\{-1, 0, +1\}$, achieving superior efficiency while preserving model performance~\cite{zafrir2019q8bert, zhu2017trained}. Notable quantization-aware training (QAT) techniques, including TWN~\cite{li2022ternaryweightnetworks}, TTQ~\cite{zhu2017trained}, and FATNN~\cite{chen2021fatnn} incorporate scaling factors to enhance ternary model optimization. The Transformer architecture, which has become essential for modern NLP tasks, has also been successfully quantized through methods such as Q8BERT~\cite{zafrir2019q8bert} and TernaryBERT~\cite{fan2020ternarybert}, demonstrating the feasibility of low-bit quantization. In computer vision, quantizing Vision Transformers (ViTs) presents unique challenges due to their sensitivity to attention distribution changes, which methods like Q-ViT~\cite{li2022qvit} address through the QAT strategy. In parallel, post-training quantization (PTQ) methods have gained significant attention for training-free advantage under relatively higher bit-width, such as 4-bit and 8-bit quantization. Representative PTQ approaches include AWQ~\cite{awq}, which identifies and preserves salient LLM weights, GPTQ~\cite{gptq} that employs layer-wise quantization with error compensation for LLMs, RepQ-ViT~\cite{Li2022RepQViTSR} that introduces scale reparameterization for vision transformers, and QwT~\cite{Fu_2025_CVPR} that achieves quantization without additional training overhead.

\textbf{Knowledge Distillation for Low-Bit.}
Pure distillation approaches such as CLIP-KD~\cite{yang2024clip} and ComKD-CLIP~\citep{chen2024comkdclip} have proven effectiveness in enhancing the performance of compressed CLIP models. Concurrently, hybrid methods combining quantization and distillation have emerged to achieve superior compression. For language models, LLM-QAT~\cite{liu2023llmqat} and BitDistiller~\cite{du2024bitdistiller} demonstrate that knowledge distillation significantly improves quantized model performance. In text embedding applications, TinyBERT~\cite{tinybert} and DistilBERT~\cite{sanh2019distilbert} show that distillation from high-precision teachers effectively bridges the accuracy gap in quantized student models.

\textbf{Multimodal Model Compression.}
As multimodal models continue to scale up, computational efficiency for edge deployment has become increasingly critical. While pruning, distillation, and quantization have been extensively studied for unimodal architectures~\cite{esser2019learned, han2016deepcompression, hinton2015distilling, jacob2018quantization, liang2021pruningquantization, zafrir2019q8bert}, ternary quantization for multimodal models remains relatively unexplored compared with unimodal counterparts such as ViT and BERT. Previous works on multimodal model compression such as TinyCLIP~\cite{wu2023tinyclip}, CLIP-KD~\cite{yang2024clip}, and ComKD-CLIP~\citep{chen2024comkdclip} employ pure distillation for parameter reduction rather than weight quantization.

% --------------------------------------------------------------------

% Methodology --------------------------------------------------------
\section{Methodology}
In this section, we present the TernaryCLIP, a framework that integrates ternarization and distillation modules to enable efficient vision-language model compression. The connection weights of TernaryCLIP are converted into a ternary format, instead of the full-precision or floating ones of the original CLIP, thereby significantly compressing the model size. Formally, we employ $\mathbf{T}$ to denote the ternary weights, where each element of $\mathbf{T}$ belongs to $\{-1,0,1\}$. Provided the input-output pair $(\boldsymbol{x}, \boldsymbol{y})$ and an apposite loss function $\mathcal{L}(\cdot,\cdot)$, we can build the following optimization in the supervised learning paradigm
\begin{equation} \label{eq:optim}
\mathbf{T}^* = \arg\min_{\mathbf{T}} \ \mathcal{L}(\boldsymbol{y},\ f(\boldsymbol{x},\mathbf{T})) \ ,   
\end{equation}
where $f(\boldsymbol{x},\mathbf{T})$ denotes the CLIP equipped with ternary weights. Intuitive approaches to solving Eq.~\eqref{eq:optim} involve converting a pre-trained model from full-precision weights into ternary weights~\citep{librecq} and computing surrogate gradients by adding a collection of full-precision weights $\mathbf{W}$ as latent variables during training~\citep{yin2018:ste}.

This work proposes an alternative approach, the key idea of which is to align the outputs of the original CLIP and our TernaryCLIP by leveraging knowledge from high-capacity teacher models. Formally, provided the loss function $\mathcal{L}(\cdot,\cdot)$ that conforms to the triangle inequality, the original optimization objective of Eq.~\eqref{eq:optim} can be relaxed by inserting $\mathbf{W}$ as follows
\begin{equation} \label{eq:triangle}
\underbrace{ \mathcal{L}(\boldsymbol{y},f(\boldsymbol{x},\mathbf{T})) }_{\text{\textcircled{\scriptsize{1}}}}
\leq
\underbrace{ \mathcal{L}(\boldsymbol{y},f(\boldsymbol{x},\mathbf{W})) }_{\text{\textcircled{\scriptsize{2}}}}
+
\underbrace{ \mathcal{L}(f(\boldsymbol{x},\mathbf{W}),f(\boldsymbol{x},\mathbf{T})) }_{\text{\textcircled{\scriptsize{3}}}} \ .   
\end{equation} 
where $\text{\textcircled{\scriptsize{1}}}$ and $\text{\textcircled{\scriptsize{2}}}$ separately describe the training losses led by ternary weight $\mathbf{T}$ and full-precision one $\mathbf{W}$, and $\text{\textcircled{\scriptsize{3}}}$ is the gap between outputs induced by using $\mathbf{W}$ and $\mathbf{T}$. Thus, the problem of solving $\text{\textcircled{\scriptsize{1}}}$ in Eq.~\eqref{eq:optim} can be implemented by that of minimizing the sum of $\text{\textcircled{\scriptsize{2}}}$ and $\text{\textcircled{\scriptsize{3}}}$. 

The new-build minimization inspires some insights; minimizing $\text{\textcircled{\scriptsize{2}}}$ implies retraining the full-precision weights $\mathbf{W}$ for pursuing high task precision, provided an obtained $\mathbf{T}$, while minimizing $\text{\textcircled{\scriptsize{3}}}$ searches for ternary weights $\mathbf{T}$ by narrowing the gap induced by using $\mathbf{W}$ and $\mathbf{T}$, provided an obtained $\mathbf{W}$. In practice, we solve this minimization by adding the ternarization-aware distillation with a straightforward ternarized module, that is, 
\begin{equation}
\label{eq:loss_ternaryclip}
\mathcal{L}_{\text{ternary}}= \mathcal{L}_{\text{task}}\left(\mathbf{W},\boldsymbol{x};\mathbf{T}\right)+\mathcal{L}_{\text{distill}}\left(\mathbf{W},\boldsymbol{x};\mathbf{T}\right) \ ,
\end{equation}
where $\mathcal{L}_{\text{task}}$ indicates contrastive loss and $\mathcal{L}_{\text{distill}}$ denotes the distillation loss that is illustrated in Figure~\ref{fig:tnclip} and decomposed as  $\mathcal{L}_{\text{distill}} = \lambda_{\text{crd}} \mathcal{L}_{\text{crd}} + \lambda_{\text{icl}} \mathcal{L}_{\text{icl}} + \lambda_{\text{fd}} \mathcal{L}_{\text{fd}}$ with $\lambda_{\text{crd}}$, $\lambda_{\text{icl}}$, and $\lambda_{\text{fd}}$ serving as balancing hyperparameters. During training, we employ ternary weights $\mathbf{T}$ for forward propagation while maintaining full-precision weights $\mathbf{W}$ for gradient updates. The task loss $\mathcal{L}_{\text{task}}$ preserves cross-modal alignment capabilities through contrastive learning, while the distillation loss $\mathcal{L}_{\text{distill}}$ transfers knowledge from the teacher model by three complementary mechanisms: contrastive relational distillation, interactive contrastive learning, and feature distillation.

\begin{algorithm}[!t]
    \caption{Ternarization-Aware Distillation}
    \label{alg:algorithm-tad}
    
    \vspace{0.3em}

    \textbf{Input:} Full-precision student and teacher CLIP model with weights $\mathbf{W}$ and $\mathbf{W}_t$, dataset $\mathcal{D}$, hyperparameter $\beta$ \\
    \textbf{Output:} Ternarized student CLIP model with weights $\mathbf{T}$ and scaling factor $\gamma$
    
    \begin{minipage}[t]{0.48\textwidth}
    \begin{enumerate}
    \item \textbf{while} not converged \textbf{do}
    \item \quad Sample minibatch $(I, T)$ from $\mathcal{D}$
    \item \quad // \textit{Ternarize weights}
    \item \quad $\gamma \leftarrow \beta\sum_{ij}|W_{ij}|/nm$
    \item \quad $\mathbf{T} \leftarrow \text{RoundClip}(\mathbf{W}/(\gamma + \epsilon), -1, 1)$
    \item \quad // \textit{Forward pass with ternarized weights}
    \item \quad $I_{e,s}, T_{e,s} \leftarrow \text{Encoders}\left(I, T; \mathbf{T}\right)$
    \item \quad $I_{e,t}, T_{e,t} \leftarrow \text{Encoders}(I, T; \mathbf{W}_t)$
    \item \quad $\mathcal{L}_{\text{ternary}} \leftarrow \mathcal{L}_{\text{task+distill}}(I_{e,s}, T_{e,s}, I_{e,t}, T_{e,t})$
    \end{enumerate}
    \end{minipage}
    \hfill
    \begin{minipage}[t]{0.48\textwidth}
    \begin{enumerate}
    \setcounter{enumi}{10}
    \item \quad // \textit{Backward pass with STE}
    \item \quad $\partial \mathcal{L}_{\text{ternary}}/\partial \mathbf{W} \leftarrow \text{STE}\left(\partial \mathcal{L}_{\text{ternary}}/\partial \mathbf{T}\right)$
    \item \quad // \textit{Update full-precision weights}
    \item \quad $\mathbf{W} \leftarrow \text{Optimizer}(\mathbf{W}, \partial \mathcal{L}_{\text{ternary}}/\partial \mathbf{W})$
    \item \textbf{end while}
    \item // \textit{Final ternarization}
    \item $\gamma \leftarrow \beta\sum_{ij}|W_{ij}|/nm$
    \item $\mathbf{T} \leftarrow \text{RoundClip}(\mathbf{W}/(\gamma + \epsilon), -1, 1)$
    \item \textbf{return} $\mathbf{T}, \gamma$
    \end{enumerate}
    \end{minipage}
\end{algorithm}

The framework of ternarization-aware distillation enables student models to learn parameters specifically optimized for ternary representation while maintaining the training stability of the full-precision format. Algorithm~\ref{alg:algorithm-tad} details the training procedure for TernaryCLIP, establishing a lightweight framework that converts CLIP models to ternary form without sacrificing performance.
A key advantage of our framework is its versatility, allowing the weights $\mathbf{W}$ to be randomly initialized for training from scratch, loaded from pre-trained models for fine-tuning, or obtained through distillation for knowledge transfer.

Here, we employ the straightforward ternarization module as follows
\[
\mathbf{T} = \text{RoundClip}\left(\frac{\mathbf{W}}{\gamma + \epsilon}, -1, 1\right) 
\quad\text{with}\quad
\gamma = \frac{\beta}{nm}\sum_{ij}|W_{ij}| \ ,
\]
where $\text{RoundClip}(x, a, b) = \max(a, \min(b, \text{round}(x)))$ constrains the rounded values within the range $[a, b]$, $\gamma$ denotes the adaptive scaling factor, $\beta$ is the hyperparameter to calibrate quantization threshold tuned by $\arg\min_{\beta} \mathcal{L}_{\text{ternary}}$, and $nm$ indicates the total number of weight elements. This scaling mechanism dynamically adjusts the ternarization threshold based on weight magnitudes, preserving the distributional characteristics of the original weights.
The process of ternary conversion involves the non-differentiable $\text{round}(\cdot)$ operation, posing a challenge for gradient-based optimization. To address this, we employ the Straight-Through Estimator (STE)~\cite{bengio2013estimating}, which enables gradient flow through the quantization function, i.e., ${\partial \mathcal{L}_{\text{ternary}}}/{\partial \mathbf{T}} \gets {\partial \mathcal{L}_{\text{ternary}}}/{\partial \mathbf{W}}$.
This module uses ternary weights $\mathbf{T}$ during forward propagation while directing gradients to the full-precision weights $\mathbf{W}$ during backpropagation.

Through joint optimization of task and distillation objectives with ternarized weights, the student model learns parameter distributions inherently suited for ternary quantization while preserving competitive performance. This joint optimization strategy ensures that the compressed model adapts to the constraints of ultra-low-bit ternary representation throughout training, rather than suffering from post-training quantization artifacts. During inference, the model operates exclusively with ternary weights, achieving significant reductions in resource consumption and computational complexity without substantial performance degradation.
% --------------------------------------------------------------------

% Experiments --------------------------------------------------------
\section{Experiments}

We trained two TernaryCLIP variants and performed extensive evaluations. Q-FFN applies ternary quantization to the embedding layer and feedforward blocks, whereas Q-ALL applies it to additional multi-head attention blocks. Table~\ref{tab:model_variants} compares the weight bit-width, quantized components, proportion, and compression ratio across the baseline CLIP model and PTQ variants, including RepQ~\citep{Li2022RepQViTSR} and QwT~\citep{Fu_2025_CVPR}. Among these models, TernaryCLIP\_Q-ALL achieves the highest quantization proportion of 99\% and the largest compression ratio of 16.98×. In the rest of the experiments, TernaryCLIP refers to Q-ALL unless specified otherwise.

\begin{table}[!htb]
\centering
\begin{tabular}{l|l|lllll|l}
\toprule
\multirow{2}{*}{\textbf{Methods}}
& \multirow{2}{*}{\textbf{Params}} & \multicolumn{5}{c|}{\textbf{Quantization}}                                                                                            & \textbf{Comp} \\ \cline{3-7}
                   &                     & \multicolumn{1}{l|}{\textbf{Weight}} & \textbf{Emb}              & \textbf{MHA}              & \multicolumn{1}{l|}{\textbf{FFN}}              & \textbf{Prop~$\uparrow$} & \textbf{Ratio~$\uparrow$}       \\ \hline
Baseline           & 149.62M             & \multicolumn{1}{l|}{32-bit}          & \texttimes & \texttimes & \multicolumn{1}{l|}{\texttimes} & 0\%                 & 1.00$\times$         \\
RepQ\_Int4         & 149.62M             & \multicolumn{1}{l|}{4-bit}           & \texttimes & \checkmark & \multicolumn{1}{l|}{\checkmark} & 82.09\%             & 3.55$\times$         \\
RepQ+QwT\_Int4     & 159.86M             & \multicolumn{1}{l|}{4-bit}           & \texttimes & \checkmark & \multicolumn{1}{l|}{\checkmark} & 76.83\%             & 3.05$\times$         \\
TernaryCLIP\_Q-FNN & 149.62M             & \multicolumn{1}{l|}{1.58-bit}        & \checkmark & \texttimes & \multicolumn{1}{l|}{\checkmark} & 71.62\%             & 3.13$\times$         \\
TernaryCLIP\_Q-ALL & 149.62M             & \multicolumn{1}{l|}{1.58-bit}        & \checkmark & \checkmark & \multicolumn{1}{l|}{\checkmark} & \textbf{99.00\%}    & \textbf{16.98$\times$} \\ \bottomrule
\end{tabular}
\caption{Comparison of quantization configuration and compression ratio for CLIP models.}
\label{tab:model_variants}
\end{table}
% Compression Ratio
% fp_size = 149.62 * 100% * 32(bit) = 4787.84
% repq_int4 = 149.62 * 0.8209 * 4(bit) + 149.62 * (1-0.8209) * 32(bit) = 491.292232+857.502144 = 1348.794376 => Compression Ratio = 1348.794376 / 4787.84 = 0.2817 = 28.17%
% repq_qwt_int4 = 159.86 * 0.7683 * 4 + 159.86 * (1-0.7683) * 32 = 1676.547736 => Compression Ratio = 1676.547776 / 5115.52 = 0.3277 = 32.77%
% tnclip_q_ffn = 149.62 * 0.7162 * 1.58 + 149.62 * (1-0.7162) * 32 = 1528.09838552 => Compression Ratio = 1528.09814 / 4787.84 = 0.3192 = 31.92%
% tnclip_q_all = 149.62 * 0.99 * 1.58 + 149.62 * (1-0.99) * 32 = 281.914004 => Compression Ratio = 281.800444 / 4787.84 = 0.0588 = 5.88%

\subsection{Experimental Setup}

\textbf{Training Configurations.} 
TernaryCLIP variants were trained on Conceptual Captions 12M (CC12M)~\citep{changpinyo2021cc12m}, a large-scale vision-language dataset, with a per-GPU batch size of 384 and a total batch size of 3072 across 8 GPUs. Table~\ref{tab:model-config} presents comprehensive configurations of CLIP models studied here, encompassing pre-trained, distilled, and quantized models. Pre-trained models are ViT-L/14 LAION, ViT-B/16 OpenAI, and LAION. For the distillation configuration, we employ LAION's CLIP ViT-L/14 as the teacher model to provide transferred knowledge, while using CLIP ViT-B/16 as the student model to achieve the balance between model performance and efficiency. Both ViT-B/16 CLIP-KD and TernaryCLIP are distilled by ViT-L/14 LAION. 4-bit PTQ models contain RepQ and RepQ+QwT quantized from ViT-B/16 OpenAI. TernaryCLIP is the 1.58-bit QAT model.

\begin{table}[!htb]
\centering
\begin{adjustbox}{max width=\textwidth}
\begin{tabular}{ll|ccccccc}
\toprule
& & \multicolumn{7}{c}{\textbf{Models}} \\
& & \textbf{CLIP LAION} & \textbf{CLIP LAION} & \textbf{CLIP OpenAI} & \textbf{CLIP-KD} & \textbf{RepQ} & \textbf{RepQ+QwT} & \textbf{TernaryCLIP} \\ \hline
\multirow{4}{*}{\rotatebox[origin=c]{90}{\textbf{Compression}}} & \textbf{Pre-Trained} & \checkmark & \checkmark & \checkmark & \texttimes & \texttimes & \texttimes & \texttimes \\
& \textbf{Quantized} & \texttimes & \texttimes & \texttimes & \texttimes & \checkmark & \checkmark & \checkmark \\
& \textbf{Distilled} & \texttimes & \texttimes & \texttimes & \checkmark & \texttimes & \texttimes & \checkmark \\
& \textbf{Parameters} & 427.6M & 149.6M & 149.6M & 149.6M & 149.6M & 159.6M & 149.6M \\ \hline
\multirow{5}{*}{\rotatebox[origin=c]{90}{\textbf{Vision}}} & \textbf{Structure} & ViT-L/14 & ViT-B/16 & ViT-B/16 & ViT-B/16 & ViT-B/16 & ViT-B/16 & ViT-B/16 \\
& \textbf{Image Size} & 224 & 224 & 224 & 224 & 224 & 224 & 224 \\
& \textbf{Patch} & 14 & 16 & 16 & 16 & 16 & 16 & 16 \\
& \textbf{Layers} & 24 & 12 & 12 & 12 & 12 & 12 & 12 \\
& \textbf{Width} & 1024 & 768 & 768 & 768 & 768 & 768 & 768 \\ \hline
\multirow{5}{*}{\rotatebox[origin=c]{90}{\textbf{Text}}} & \textbf{Vocab Size} & 49408 & 49408 & 49408 & 49408 & 49408 & 49408 & 49408 \\
& \textbf{Context} & 77 & 77 & 77 & 77 & 77 & 77 & 77 \\
& \textbf{Layers} & 12 & 12 & 12 & 12 & 12 & 12 & 12 \\
& \textbf{Width} & 768 & 512 & 512 & 512 & 512 & 512 & 512 \\
& \textbf{Heads} & 12 & 12 & 12 & 12 & 12 & 12 & 12 \\ \hline
& \textbf{Embed\_Dim} & 768 & 512 & 512 & 512 & 512 & 512 & 512 \\ \bottomrule
\end{tabular}
\end{adjustbox}
\caption{Model configurations, including training strategy (whether to use distillation or quantization), number of parameters, embedding dimension, vision, and text configurations.}
\label{tab:model-config}
\end{table}

\textbf{Hyperparameter Tuning.} 
Systematic hyperparameter tuning was conducted through a two-stage process. First, we performed exploratory training for 32 epochs to identify the selection of promising hyperparameters and assess convergence behavior. Subsequently, we trained the final model for 64 epochs using the AdamW optimizer with these tuned hyperparameters, learning rate of $1 \times 10^{-3}$, $1 \times 10^{4}$ warm-up steps, cosine annealing schedule, and weight decay of 0.1. The distillation loss factors are determined as $\lambda_{\text{crd}}=1.0$, $\lambda_{\text{icl}}=1.0$, and $\lambda_{\text{fd}}=2000.0$ to balance different knowledge transfer strategies. Hyperparameter configurations are detailed in Appendix~\ref{sec:hyperparam-tuning}.

\textbf{Evaluation Configurations.} 
Comprehensive evaluations were conducted across 41 diverse datasets to assess both effectiveness and efficiency. The evaluations comprise 37 single-label image classification datasets, 1 multi-label image classification dataset, and 3 image-text retrieval datasets. All experiments follow a zero-shot evaluation protocol to demonstrate the generalization capabilities of CLIP models. Performance metrics include Accuracy@1 for single-label image classification, Mean Average Precision (mAP) for multi-label image classification, and Recall@5 for image-text retrieval. To evaluate inference efficiency, we benchmarked compatible models with various precision representations on resource-constrained hardware (Apple M4 Pro ARM CPU), measuring crucial deployment metrics including sparsity, storage footprint, memory consumption, and inference latency, which simulates real-world edge deployment scenarios. Ablation studies examining data augmentation, activation quantization, and self-distillation strategies are presented in Appendix~\ref{sec:ablation-study}.

\textbf{Baseline Comparisons.} 
We compare TernaryCLIP against multiple baselines to demonstrate its effectiveness. LAION CLIP ViT-L/14 serves as the teacher model for both CLIP-KD and our TernaryCLIP variants, enabling the comparison of the distillation baseline. We also include OpenAI and LAION CLIP ViT-B/16 pre-trained models as full-precision baselines for evaluating compressed variants. Additionally, we employ two 4-bit PTQ methods, RepQ~\citep{Li2022RepQViTSR} and RepQ+QwT~\citep{Fu_2025_CVPR}, on OpenAI CLIP ViT-B/16 as quantization baselines to compare our approach against existing post-training quantization techniques.

\textbf{Inference Configurations.} 
To ensure reproducibility and fair comparison, we standardize the inference pipeline across all compatible CLIP models. We adopt the \href{https://github.com/ggml-org/ggml/blob/master/docs/gguf.md}{GGUF} for model storage and leverage the \href{https://github.com/ggml-org/ggml}{GGML} framework for efficient inference. For the TernaryCLIP, we implement the TQ1\_0 quantization format specifically designed for ternary weights, which enables efficient bit-packing, unpacking, and matrix multiplication operations that minimize both memory overhead and computational latency. Alternative quantization schemes, such as Q4\_0, are evaluated in Appendix~\ref{sec:latency}. Each benchmark measurement represents the average value of 1,000 independent runs to ensure statistical reliability, with results reported in Table~\ref{tab:model-sparsity-storage-mem-latency}.

\subsection{Zero-Shot Image Classification Performance}
\label{section:zero-shot-classification}
\textbf{Classification Data Types.} The evaluation of TernaryCLIP, alongside other models, involves 37 image classification datasets categorized into three distinct types: 21 natural, 7 specialized, and 9 structured. Natural datasets consist of images depicting real-world objects and scenes encountered by humans, such as those in the \textit{ImageNet series}~\citep{deng2009imagenet}. Specialized datasets are tailored toward domain-specific applications that require expert knowledge, such as those in the \textit{PatchCamelyon (PCam)} that contains histopathological scans of lymph node sections~\citep{veeling2018rotation}. Structured datasets focus on spatial relationships, such as \textit{CLEVR}, a synthetic visual question answering dataset~\citep{johnson2017clevr}. Comprehensive dataset-specific performance results are presented in Appendix~\ref{sec:classification}.

\begin{figure}[!htb]
    \centering
    \begin{minipage}[b]{0.48\textwidth}
        \centering
        \includegraphics[width=\textwidth]{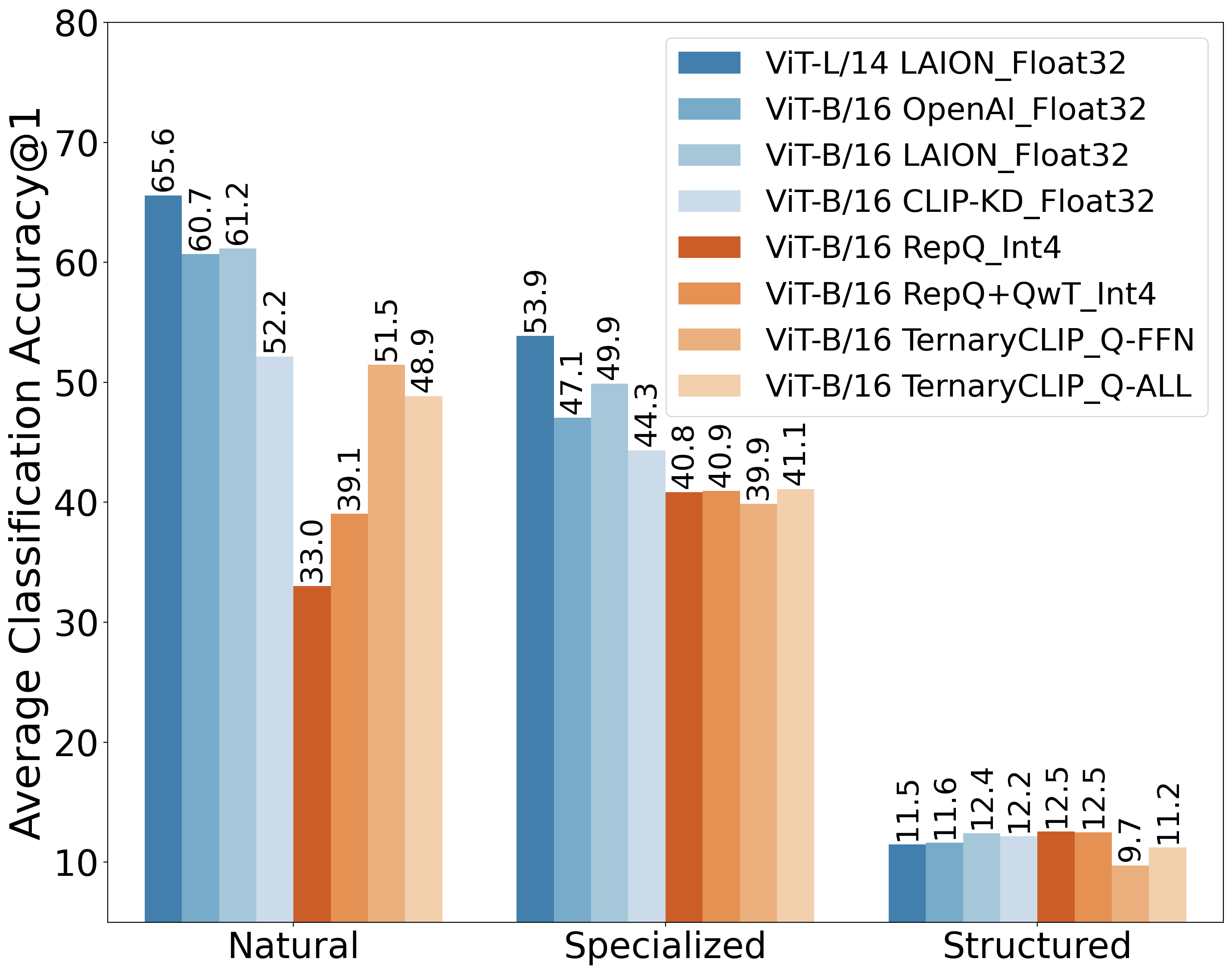}
        \caption{Zero-shot image classification performance (Accuracy@1) on three dataset types and CLIP models. ViT-L/14 or ViT-B/16 indicates which image encoder is used for the structure of CLIP.}
        \label{fig:classification-ds-types}
    \end{minipage}
    \hfill
    \begin{minipage}[b]{0.48\textwidth}
        \centering
        \includegraphics[width=\textwidth]{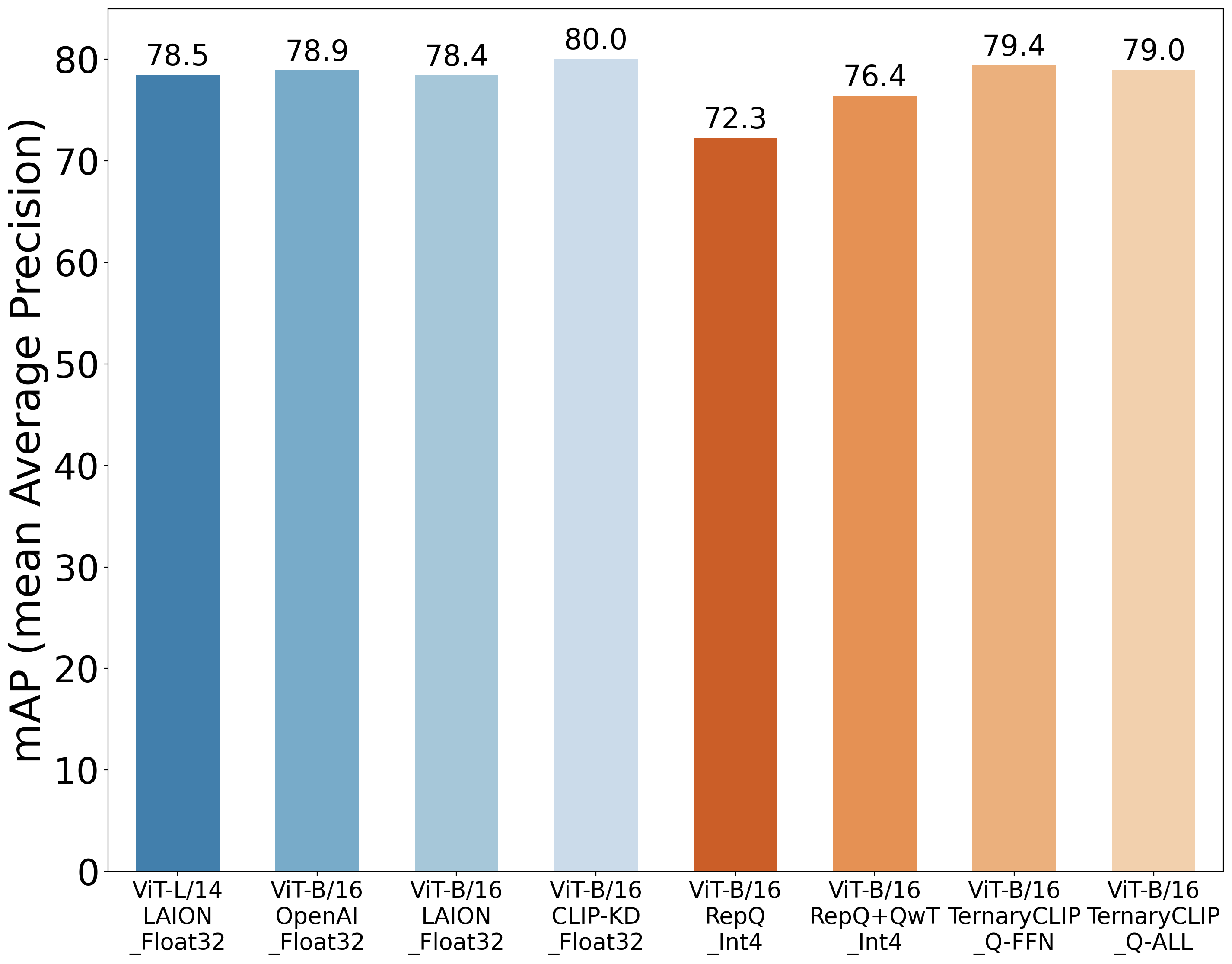}
        \caption{Zero-shot multi-label classification performance (mean average precision) on PASCAL VOC 2007. ViT-L/14 or ViT-B/16 indicates which image encoder is used for the structure of CLIP.}
        \label{fig:multi-label-classification}
    \end{minipage}
\end{figure}

\begin{table}[!htb]
\centering
\begin{tabular}{l|llll|l}
\toprule
\textbf{Models}              & \textbf{Pre-Trained} & \textbf{Quantized} & \textbf{Distilled} & \textbf{Weight} & \textbf{Average~$\uparrow$} \\ \hline
ViT-L/14 LAION              & \checkmark                  & \texttimes                 & \texttimes                 & 32-bit          & 50.20\%          \\ \hline
ViT-B/16 OpenAI             & \checkmark                  & \texttimes                 & \texttimes                 & 32-bit          & 46.17\%          \\ \hline
ViT-B/16 LAION              & \checkmark                  & \texttimes                 & \texttimes                 & 32-bit          & 47.18\%          \\ \hline
ViT-B/16 CLIP-KD            & \texttimes                   & \texttimes                 & \checkmark                & 32-bit          & 40.95\%          \\ \hline
ViT-B/16 RepQ               & \texttimes                   & \checkmark                & \texttimes                 & 4-bit           & 29.52\%          \\ \hline
ViT-B/16 RepQ+QwT           & \texttimes                   & \checkmark                & \texttimes                 & 4-bit+32-bit    & 32.95\%          \\ \hline
ViT-B/16 TernaryCLIP\_Q-FFN & \texttimes                   & \checkmark                & \checkmark                & 1.58-bit        & 39.12\%          \\ \hline
ViT-B/16 TernaryCLIP\_Q-ALL & \texttimes                   & \checkmark                & \checkmark                & 1.58-bit        & 38.24\%          \\ \bottomrule
\end{tabular}
\caption{Zero-shot image classification average performance (Accuracy@1) on 37 datasets.}
\label{tab:zero-shot-classification}
\end{table}

\textbf{Evaluation of Classification.} 
Figure~\ref{fig:classification-ds-types} and Figure~\ref{fig:multi-label-classification} demonstrate the effectiveness of TernaryCLIP in preserving performance under weight quantization. The analysis reveals four key findings.
First, the performance degradation scales with the full-precision baseline model capacity. Specifically, TernaryCLIP exhibits an average performance decrease of 8.43\% relative to the same-sized ViT-B/16 OpenAI and LAION, and 11.96\% relative to the larger teacher model ViT-L/14 LAION, both of which are trained from scratch in Figure~\ref{fig:classification-ds-types} and Table~\ref{tab:zero-shot-classification}. Despite this degradation, the substantial compression ratio of 16.98× makes TernaryCLIP well-suited for edge devices.
Second, distillation provides consistent benefits across CLIP variants. Both CLIP-KD and TernaryCLIP achieve over 53\% Accuracy@1 on ImageNet-1K~\citep{yang2024clip}, resulting in a 17\% improvement over the baseline model. This validates the effectiveness of the distillation module in TernaryCLIP.
Third, on the multi-label classification dataset PASCAL VOC 2007 in Figure~\ref{fig:multi-label-classification}, TernaryCLIP variants maintain mAP scores within 1\% of full-precision models while outperforming the best PTQ method by 2.6\%.
Last, TernaryCLIP obtains a remarkable compression ratio with promising accuracy. Compared with the full-precision CLIP-KD, our approach incurs only controllable accuracy loss. Meanwhile, it significantly outperforms 4-bit PTQ methods, including RepQ and RepQ+QwT. In Table~\ref{tab:model_variants} and Table~\ref{tab:zero-shot-classification}, TernaryCLIP\_Q-ALL achieves a 16.98× compression ratio using 1.58-bit weights, with only 2.7\% accuracy drop compared with the 32-bit CLIP-KD baseline. Notably, it outperforms RepQ+QwT by 5.29\%, despite the latter using higher bit-width representation, 4-bit weights with 32-bit compensation, providing only 3.13× compression ratio. These results demonstrate that TernaryCLIP achieves superior compression and competitive performance by integrating ternarization and distillation.

\begin{table}[!h]
\centering
\begin{adjustbox}{max width=\textwidth}
\begin{tabular}{l|lllll}
\toprule
 \textbf{Models} & \textbf{Precision} & \textbf{Sparsity}~$\uparrow$ & \textbf{Storage (MB)}~$\downarrow$ & \textbf{Memory (MB)}~$\downarrow$ & \textbf{Latency (ms)}~$\downarrow$ \\
\hline
CLIP & Float32 & 0\% & 1630.90 (100\%) & 1690.27 (100\%) & 886.15 (100\%) \\
ViT-L/14 & Float16 & 0\% & 817.28 (50\%) & 876.65 (52\%) & 710.88 (80\%) \\
\hline
CLIP & Float32 & 0\% & 571.60 (35\%) & 593.76 (35\%) & 241.40 (27\%) \\
ViT-B/16 & Float16 & 0\% & 287.73 (18\%) & 309.90 (18\%) & 174.62 (20\%) \\
\hline
TernaryCLIP\_Q-FFN & \multirow{2}{*}{TQ1\_0} & \multirow{2}{*}{44.57\% (62.27\%)} & \multirow{2}{*}{105.00 (6\%)} & \multirow{2}{*}{127.18 (8\%)} & \multirow{2}{*}{126.05 (14\%)} \\
ViT-B/16 & & & & & \\
\hline
TernaryCLIP\_Q-ALL & \multirow{2}{*}{TQ1\_0} & \multirow{2}{*}{\textbf{60.88\% (61.56\%)}} & \multirow{2}{*}{\textbf{35.25 (2\%)}} & \multirow{2}{*}{\textbf{57.45 (3\%)}} & \multirow{2}{*}{\textbf{106.75 (12\%)}} \\
ViT-B/16 & & & & & \\
\bottomrule
\end{tabular}
\end{adjustbox}
\caption{CLIP model sparsity, storage, memory, and latency on different precisions. For the sparsity x\% (y\%), x\% represents the total sparsity of all parameters, and y\% represents the partial sparsity of all ternary parameters.}
\label{tab:model-sparsity-storage-mem-latency}
\end{table}

\textbf{Evaluation of Efficiency.} 
Table~\ref{tab:model_variants} presents the efficiency metrics of quantization proportion and compression ratio for quantized models, while TernaryCLIP achieves the highest 99\% proportion and 16.98× compression ratio.
Table~\ref{tab:model-sparsity-storage-mem-latency} presents significant efficiency improvements achieved by TernaryCLIP in terms of sparsity, storage, memory, and latency. TernaryCLIP\_Q-ALL achieves 60.88\% weight sparsity, enabling further optimization through sparse matrix operations and creating opportunities for additional compression techniques. The ternary weight distributions for sparsity illustration are detailed in Appendix~\ref{sec:ternary-weight-distribution}. The most compressed variant of TernaryCLIP requires only 35.25MB storage, a dramatic reduction from 1630.90MB for the ViT-L/14 Float32 teacher model and 571.60MB for the ViT-B/16 Float32 baseline model, representing 46× and 16× reductions. This enables deployment on storage-constrained edge devices to host CLIP models. Inference-time memory consumption of model weights and intermediate computations drops from 1690.27MB for the teacher model and 593.76MB for the baseline model to only 57.45MB, achieving 29× and 10× reductions. This breakthrough enables inference with limited RAM on edge devices. TernaryCLIP reduces latency from 886.15ms for the teacher model and 241.40ms for the baseline model to only 106.75ms, delivering 8.3× and 2.3× latency improvements. This acceleration enables real-time applications with strict latency constraints.

TernaryCLIP obtains competitive zero-shot classification performance with aggressive 1.58-bit weight quantization, representing a huge compression ratio compared with full-precision and post-training quantized models, while achieving substantial reductions in storage, memory, and inference costs. These results validate our approach as a practical solution for deploying multimodal models on resource-constrained devices.

\begin{figure}[!h]
    \centering
    \begin{minipage}[b]{0.48\textwidth}
        \centering
        \includegraphics[width=\textwidth]{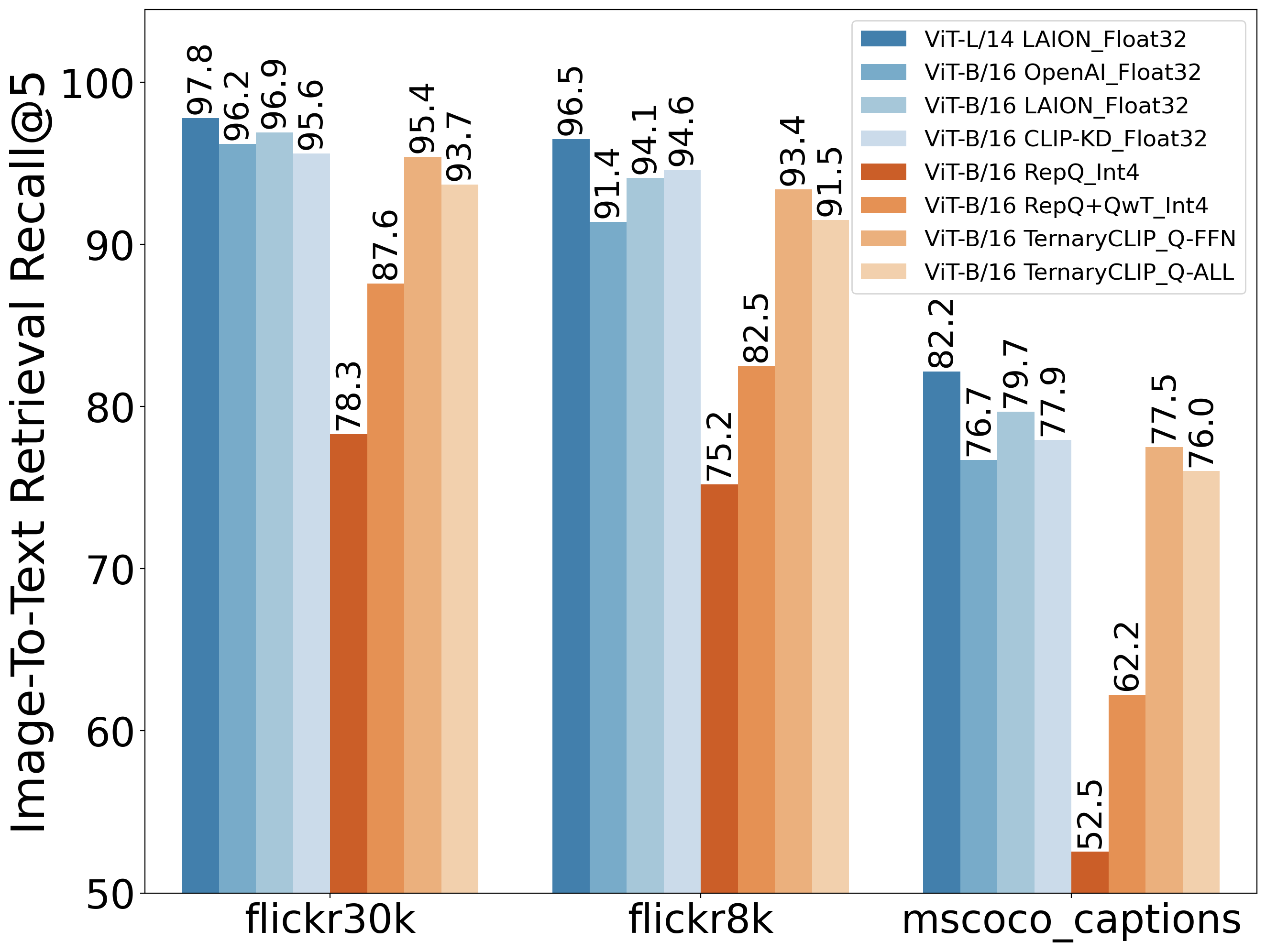}
        \caption{Zero-shot image-to-text retrieval performance (Recall@5) on three datasets and CLIP models. ViT-L/14 or ViT-B/16 indicates which image encoder is used for the structure of CLIP.}
        \label{fig:i2t}
    \end{minipage}
    \hfill
    \begin{minipage}[b]{0.48\textwidth}
        \centering
        \includegraphics[width=\textwidth]{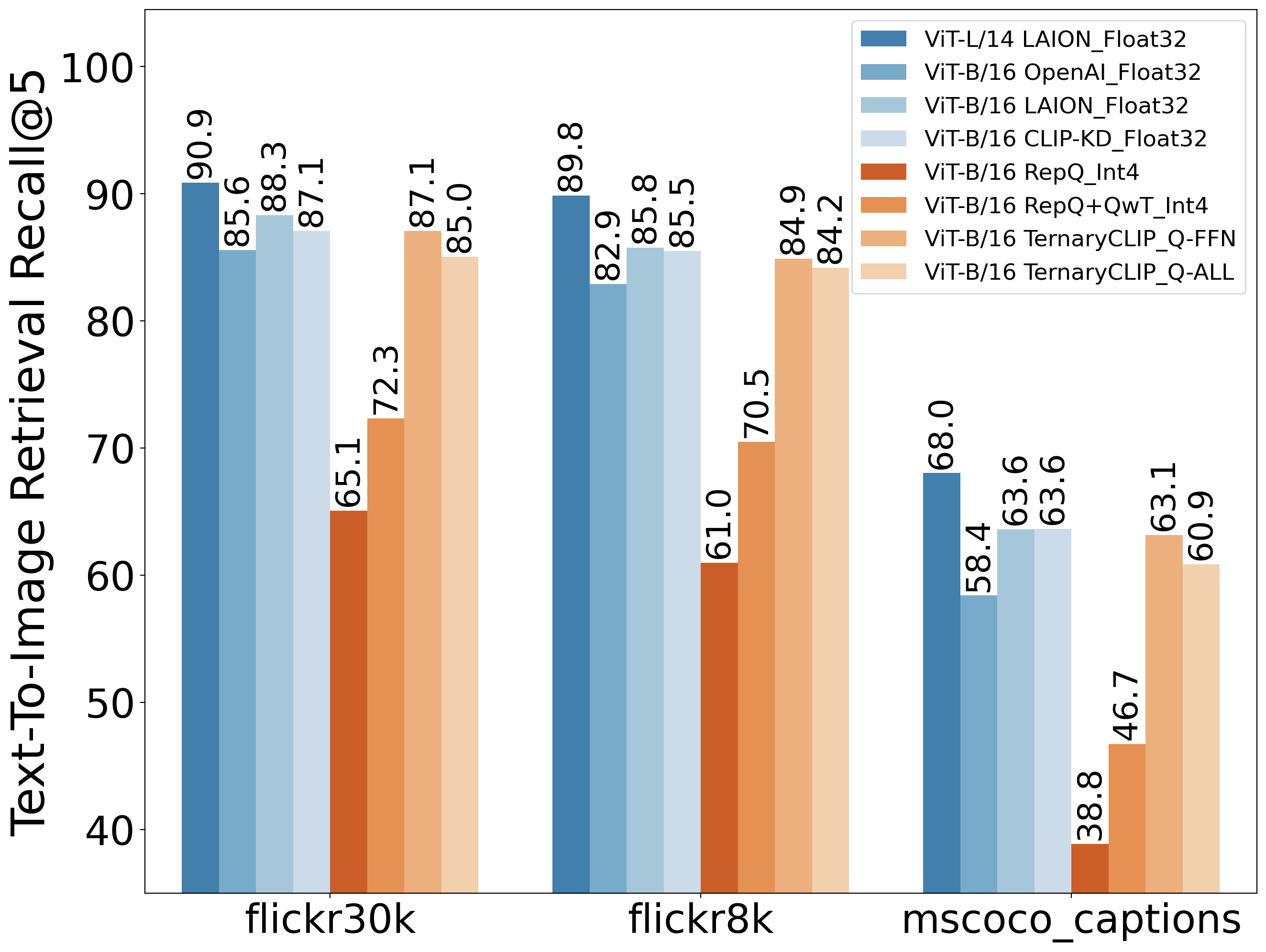}
        \caption{Zero-shot text-to-image retrieval performance (Recall@5) on three datasets and CLIP models. ViT-L/14 or ViT-B/16 indicates which image encoder is used for the structure of CLIP.}
        \label{fig:t2i}
    \end{minipage}
\end{figure}

\subsection{Zero-Shot Image-Text Retrieval Performance}

\textbf{Datasets.} To assess zero-shot cross-modal retrieval performance, we select three widely adopted datasets, including Flickr8k~\cite{hodosh2013framing}, Flickr30k~\cite{young2014image}, and MS COCO~\cite{lin2014microsoft}. We evaluated two retrieval tasks on these datasets, including image-to-text and text-to-image. Performance is measured using Recall@5 as the evaluation metric.

\textbf{Evaluation.} As shown in Figure~\ref{fig:i2t} and Figure~\ref{fig:t2i}, we conducted comprehensive evaluations on various CLIP variants.
When evaluated against full-precision pre-trained models (ViT-L/14 LAION, ViT-B/16 OpenAI, and LAION), TernaryCLIP exhibits competitive performance with less than 3\% degradation compared with models of equivalent size, and approximately 5\% degradation compared with the larger model.
Compared with the distillation-only CLIP-KD model under similar training configurations, TernaryCLIP incurs an average performance degradation of only 2.17\%.
Notably, TernaryCLIP significantly outperforms 4-bit PTQ methods, including RepQ and RepQ+QwT. Leveraging the QAT module, TernaryCLIP with merely 1.58-bit weights achieves an average performance improvement of 11.5\% over the best PTQ model, RepQ+QwT employing 4-bit quantized weights and 32-bit compensation weights. These results demonstrate that our approach achieves both the lowest bit-width representation and the highest retrieval performance compared with PTQ methods. Beyond this favorable trade-off between performance and compression, the integration of ternarization-aware training and distillation obtains substantial efficiency improvements, as detailed in Section~\ref{section:zero-shot-classification}.

\subsection{Covergence and Costs}

\begin{figure}[!htb]
    \centering
    \includegraphics[width=0.6\linewidth]{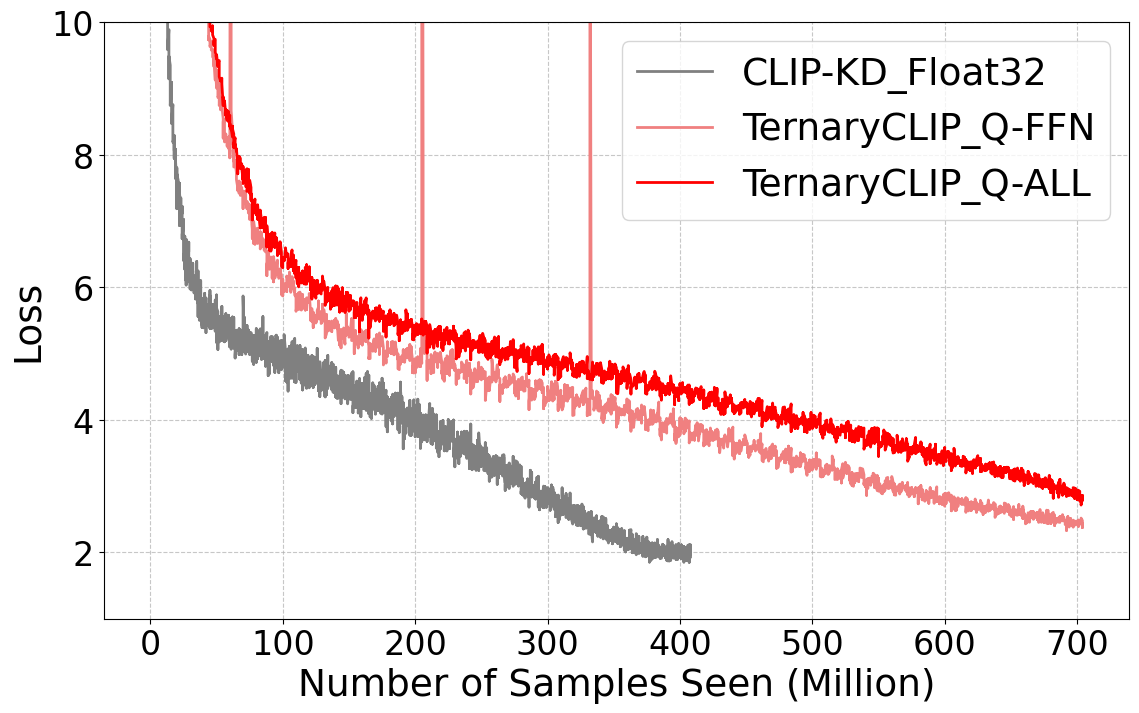}
    \caption{Training total loss curves of models on the number of samples seen: 1) full-precision CLIP-KD, 2) TernaryCLIP\_Q-FFN with ternarized EMB+FFN, 3) TernaryCLIP\_Q-ALL with ternarized EMB+MHA+FFN.}
    \label{fig:training-losses}
\end{figure}

% \subsection{Training Process of TnCLIP}
\textbf{Convergence.} 
Analysis of the training loss curves in Figure~\ref{fig:training-losses} reveals two critical insights of training ternary models. The constituent losses including $\mathcal{L}_{\text{task}}$, $\mathcal{L}_{\text{crd}}$, $\mathcal{L}_{\text{icl}}$, and $\mathcal{L}_{\text{fd}}$ are detailed in Appendix~\ref{sec:train-tnclip}.
First, ternary models exhibit oscillatory convergence behavior, in contrast to the smooth trajectory of full-precision training. This distinction arises from the fundamental difference between continuous and discrete parameter spaces. While full-precision models optimize over a continuous weight space, ternary models operate within a discrete weight space defined by the quantization function $\text{RoundClip}(\cdot)$. Consequently, gradient updates that induce minor changes in full-precision weights can trigger abrupt transitions in ternary weights, resulting in periodic fluctuations in the training loss curves.
Second, the quantization proportion directly impacts optimization difficulty. As the proportion of ternarized structures increases, the accumulated quantization error intensifies, requiring a prolonged training budget to achieve convergence and recover from performance degradation. Experimental results demonstrate that, given a sufficient training budget, ternary models can approach the performance of their full-precision counterparts, indicating the viability of ultra-low-bit quantization. This finding suggests that the performance upper bound of ternary models is constrained primarily by training resources rather than the inherent architectural limitation imposed by the compression technique, quantization.

\textbf{Scaling Law for Ternary Quantization.}
Inspired by scaling laws in language models~\citep{kaplan2020scaling}, we propose a scaling law that characterizes the relationship between a training budget and quantization-induced performance degradation. This law provides a predictive framework for understanding how increased training can mitigate quantization error. Specifically, we formulate the performance gap as
\begin{equation*}
    \Delta(C,q) = \alpha C^{-\beta} ~ f(q) + c = \alpha C^{-\beta} ~ (1-q)^{-\gamma} + c \ , 
    \quad \text{s.t.}\quad \alpha, \beta, \gamma > 0 \ ,
\end{equation*}
where $\Delta(C,q)$ represents the performance gap between the quantized and full-precision models, $C$ denotes the training budget, $\alpha$ and $\beta$ are power law parameters characterizing the compute-performance scaling, $f(q)$ is the quantization penalty function with the quantized proportion $q \in [0,1]$, $\gamma$ controls the sensitivity of quantization extent influencing performance degradation, and $c$ is the regularization constant.

\begin{table*}[!htb]
\centering
\begin{tabular}{l|llll}
\toprule
\textbf{Models} & \textbf{Datasets} & \textbf{GPU} & \textbf{Time}~$\downarrow$ & \textbf{Price}~$\downarrow$ \\
\hline
ViT-L/14 LAION & LAION-400M & 400 * A100 & 127 hours & 44,704\$ \\
\hline
ViT-L/14 OpenAI & WIT-400M & 256 * V100 & 288 hours & 34,818\$ \\
\hline
ViT-B/16 LAION & LAION-400M & 176 * A100 & 61 hours & 9,448\$ \\
\hline
ViT-B/16 CLIP-KD & CC3M+CC12M & 8 * A800 & 137 hours & 998\$ \\
\hline
ViT-B/16 TernaryCLIP & CC12M & 8 * 6000Ada & 129 hours & \textbf{575\$} \\
\bottomrule
\end{tabular}
\caption{Comparison of CLIP model training specifications, including Vision Transformer architecture, dataset, GPU requirements, training time, and associated approximate costs.}
\label{tab:clip_training_costs}
\end{table*}

\textbf{Training Cost Analysis.}
To contextualize the computational efficiency, we utilize training resources to compare TernaryCLIP against baseline CLIP models, including OpenAI~\citep{radford2021learning}, \href{https://laion.ai/}{LAION}, and CLIP-KD~\citep{yang2024clip}. While OpenAI does not disclose the detailed training configuration for the ViT-B/16 model, detailed specifications are available for LAION and CLIP-KD, facilitating quantitative comparison. 
Table~\ref{tab:clip_training_costs} summarizes the GPU-hours and estimated costs for each model, where costs are derived from average pricing on commercial cloud platforms. The results reveal apparent differences in computational requirements across training paradigms. Full-precision models trained from scratch, such as LAION and OpenAI, require extensive GPU resources.
In contrast, distillation methods like CLIP-KD and TernaryCLIP achieve competitive performance with significantly reduced training costs. Moreover, TernaryCLIP achieves additional efficiency through ternary quantization.

Overall, the performance trade-off is acceptable given the efficiency gains achieved by TernaryCLIP.
Experimental results reveal that under identical model architecture and training configurations, the performance degradation from full-precision CLIP to ternary CLIP is controllable, averaging only 2.7\% as shown in Table~\ref{tab:zero-shot-classification}. Meanwhile, TernaryCLIP demonstrates significant improvements in efficiency metrics, including quantization proportion, compression ratio, sparsity, storage, memory, and latency in Table~\ref{tab:model_variants} and Table~\ref{tab:model-sparsity-storage-mem-latency}.
These results demonstrate that our TernaryCLIP achieves a great balance between performance preservation and computational efficiency, making it particularly well-suited for deployment in resource-constrained environments such as edge devices.
% 量化蒸馏和单纯蒸馏的模型在其他因素相同的情况下的性能差距极小，说明量化蒸馏算法的优越性和推理Efficiency很好。这种微小性能损失+推理效率提升，很符合端侧设备部署模型的需求
% Conclusion ---------------------------------------------------------
\zhangsqchecked{\section{Conclusions and Prospects}}
% Conclusions
In this paper, we propose the TernaryCLIP, which compresses both the vision and text encoders of CLIP into the ternary format to meet the demands for reducing memory consumption, storage footprint, inference latency, and annotation costs for downstream tasks.
By integrating ternarization and distillation modules, TernaryCLIP successfully compresses model parameters to an ultra-low 1.58-bit weight precision while achieving up to 99\% ternary weight, 16.98× compression ratio, 2.3× inference acceleration, 16× storage reduction, 10× memory optimization, and 60\% sparsity without compromising the model's zero-shot capabilities evaluated on comprehensive cross-modal understanding tasks.
Experimental results across 41 benchmarks demonstrate the effectiveness and efficiency of TernaryCLIP, supporting deployment on resource-constrained edge devices. 
Therefore, TernaryCLIP establishes a new paradigm for lightweight vision-language models that maintains competitive performance with substantially reduced resource requirements, providing a practical solution for deploying large-scale multimodal models.

% Prospects
There are several avenues for future work that warrant investigation.
Firstly, ternarization-aware distillation from pre-trained models reduces the student model's ability to adapt to domain-specific applications. Retraining is still necessary to achieve better performance on specific tasks.
Secondly, quantization introduces information loss, which can hinder the model's ability to capture subtle image-text alignments compared with full-precision models trained from scratch.
Thirdly, the current implementation only quantizes weights while leaving activations in FP16 format. Future work could explore activation quantization, such as INT8 and specialized hardware implementations that leverage bit-wise operations for ternary computations, potentially achieving further efficiency gains.
Lastly, this work focuses on image-text alignment tasks. The extensions of ternary quantization techniques to other modalities or architectures remain unexplored and present opportunities for future research.

\section*{Acknowledgments and Disclosure of Funding}
Shao-Qun Zhang is the corresponding author, with email \textit{zhangsq@lamda.nju.edu.cn}. This research was supported by the Jiangsu Provincial Natural Science Foundation Youth Project (BK20230782).

% Bibliography -------------------------------------------------------
\bibliography{JMref}
\bibliographystyle{plain}
% \bibliographystyle{alpha}
% plain, 按字母的顺序排列, 比较次序为作者-年度-标题
% unsrt, 样式同plain, 只是按照引用的先后排序
% alpha, 用作者名首字母+年份后两位作标号, 以字母顺序排序
% abbrv, 类似plain, 将月份全拼改为缩写, 更显紧凑
% ieeetr, 国际电气电子工程师协会期刊样式
% acm, 美国计算机学会期刊样式
% siam, 美国工业和应用数学学会期刊样式
% apalike, 美国心理学学会期刊样式

% --------------------------------------------------------------------

% Appendix -----------------------------------------------------------
\appendix
\section{Training Hyperparameters}
\label{sec:hyperparam-tuning}
% Appendix B. 绘制一个table 展示所有的超参+粗体为选择的超参数
In Table~\ref{tab:model-hyperparam}, we exhibit hyperparameters used to train our TernaryCLIP model, while the values between square brackets are the to-be-selected hyperparameters and the values with the text bold font are the selected hyperparameters after the tuning procedure.

\begin{table}[!ht]
\centering
\begin{adjustbox}{max width=\textwidth}
\begin{tabular}{l|l}
\hline
\textbf{Hardware} & 8 * NVIDIA RTX 6000 Ada \\
\hline
\textbf{Model} & Teacher: [\textbf{ViT-L/14}] \\
 & Student: [\textbf{ViT-B/16}] \\
\hline
\textbf{Weight} & [OpenAI\_400M\_ep32, \textbf{Laion400M\_ep32}] \\
\hline
\textbf{Quantization} & weight\_quant: [\textbf{ternary}, int3, int4] \\
 & activation\_quant: [int8, \textbf{float16}] \\
\hline
\textbf{Ternarization} & $\beta$: [1, \textbf{2}, 3], $\epsilon$: [\textbf{1e-6}] \\
\hline
\textbf{Precision} & [\textbf{amp}, amp\_bf16, bp16, fp32] \\
\hline
\textbf{Data Load} & num\_workers: [4, 8, \textbf{16}, 32] \\
\hline
\textbf{Epoch} & [32, \textbf{64}] \\
\hline
\textbf{Learning Rate} & [1e-4, 5e-4, \textbf{1e-3}] \\
\hline
\textbf{Warmup} & [1000, 5000, \textbf{10000}, 100000] \\
\hline
\textbf{Batch Size} & [128$\times$8, 256$\times$8, \textbf{384}$\times$8, 512$\times$8] \\
\hline
\textbf{Optimizer} & adamw(\textbf{0.9}, [0.98, 0.998, \textbf{0.999}], $\epsilon$: [\textbf{1e-6}]) \\
\hline
\textbf{Weight Decay} & [0.01, 0.05, \textbf{0.1}, 0.2] \\
\hline
$\boldsymbol{\lambda_{\text{crd}},\tau_{crd}}$ & [0.5, \textbf{1}, 2], [0.5, \textbf{1}, 2] \\
\hline
$\boldsymbol{\lambda_{\text{icl}}}$ & [0.5, \textbf{1}, 2] \\
\hline
$\boldsymbol{\lambda_{fd}}$ & [1000, \textbf{2000}, 4000] \\
\hline
\textbf{Augment Config} & [\textbf{None}, Scale, Scale+Color\_Jitter+Gray\_Scale] \\
\hline
\end{tabular}
\end{adjustbox}
\caption{Training hyperparameters of TernaryCLIP: the value with bold font is the recommended hyperparameters after tuning. For example, $\lambda_{\text{crd}}=1.0$ and temperature $\tau_{\text{crd}}=1.0$.}
\label{tab:model-hyperparam}
\end{table}

\clearpage

% Details of Training TernaryCLIP ---------------------------------------------
\section{Details of Training TernaryCLIP}
\label{sec:train-tnclip}
% Appendix C. 把所有loss的组成的图像绘制出来 loss curves + valid curve

\begin{figure}[!ht]
    \centering
    \begin{minipage}[b]{0.48\textwidth}
        \centering
        \includegraphics[width=\textwidth]{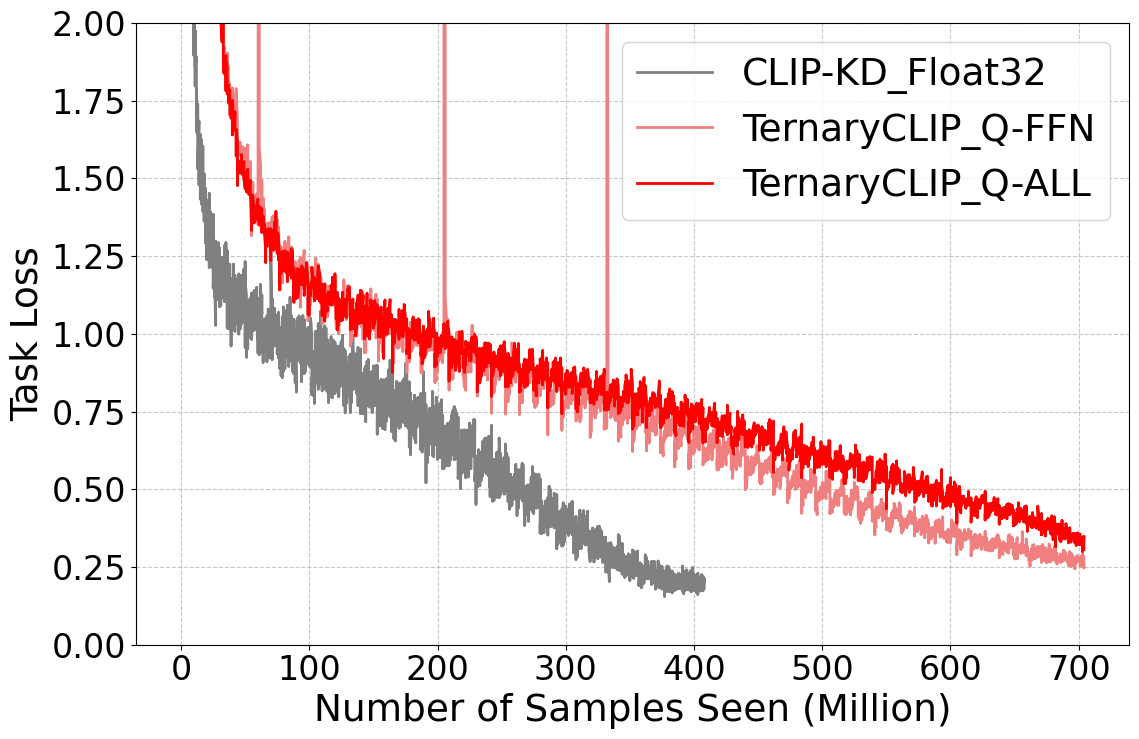}
        \caption{Task losses of full-precision CLIP-KD, TernaryCLIP\_Q-FFN and TernaryCLIP\_Q-ALL.}
        \label{fig:task-losses}
    \end{minipage}
    \hfill
    \begin{minipage}[b]{0.48\textwidth}
        \centering
        \includegraphics[width=\textwidth]{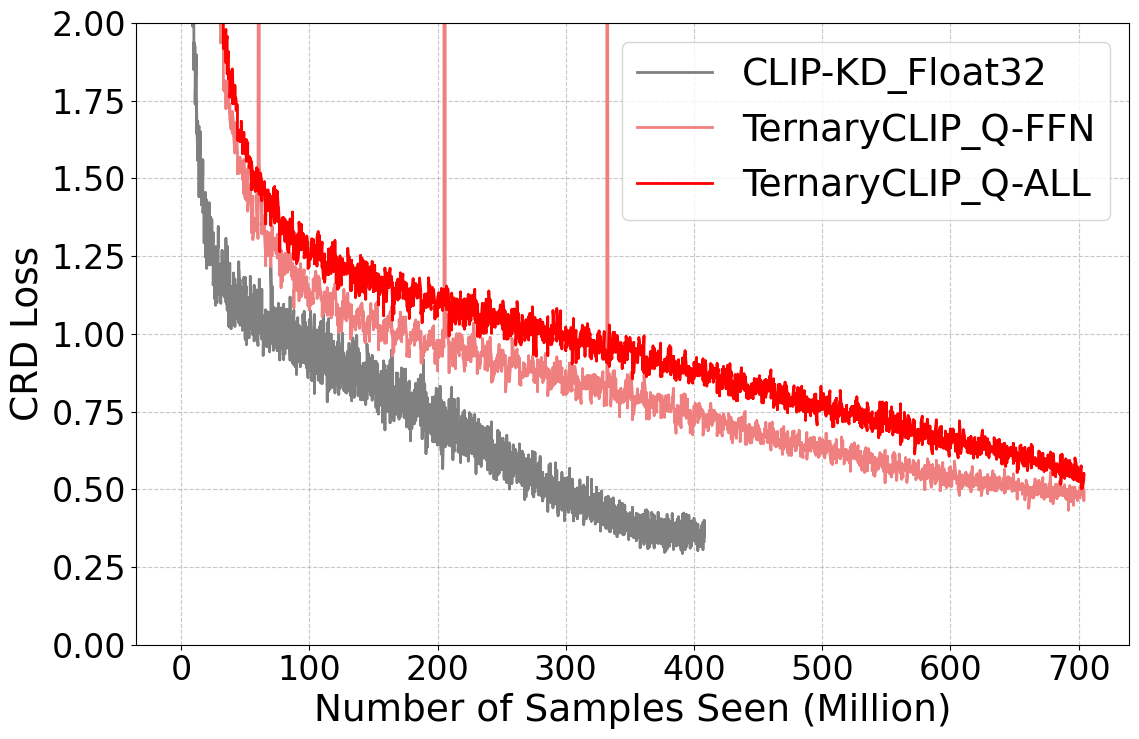}
        \caption{CRD losses of full-precision CLIP-KD, TernaryCLIP\_Q-FFN and TernaryCLIP\_Q-ALL.}
        \label{fig:crd-losses}
    \end{minipage}
    
    \vspace{0.5cm}
    
    \begin{minipage}[b]{0.48\textwidth}
        \centering
        \includegraphics[width=\textwidth]{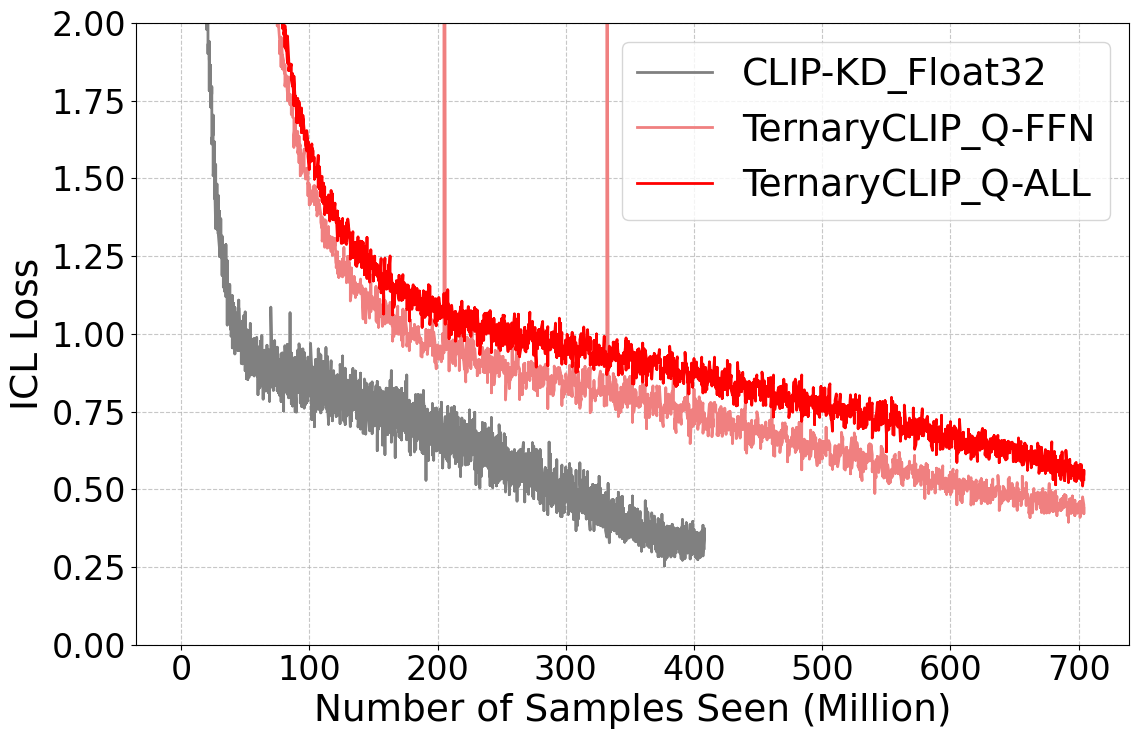}
        \caption{ICL losses of full-precision CLIP-KD,  TernaryCLIP\_Q-FFN and TernaryCLIP\_Q-ALL.}
        \label{fig:icl-losses}
    \end{minipage}
    \hfill
    \begin{minipage}[b]{0.48\textwidth}
        \centering
        \includegraphics[width=\textwidth]{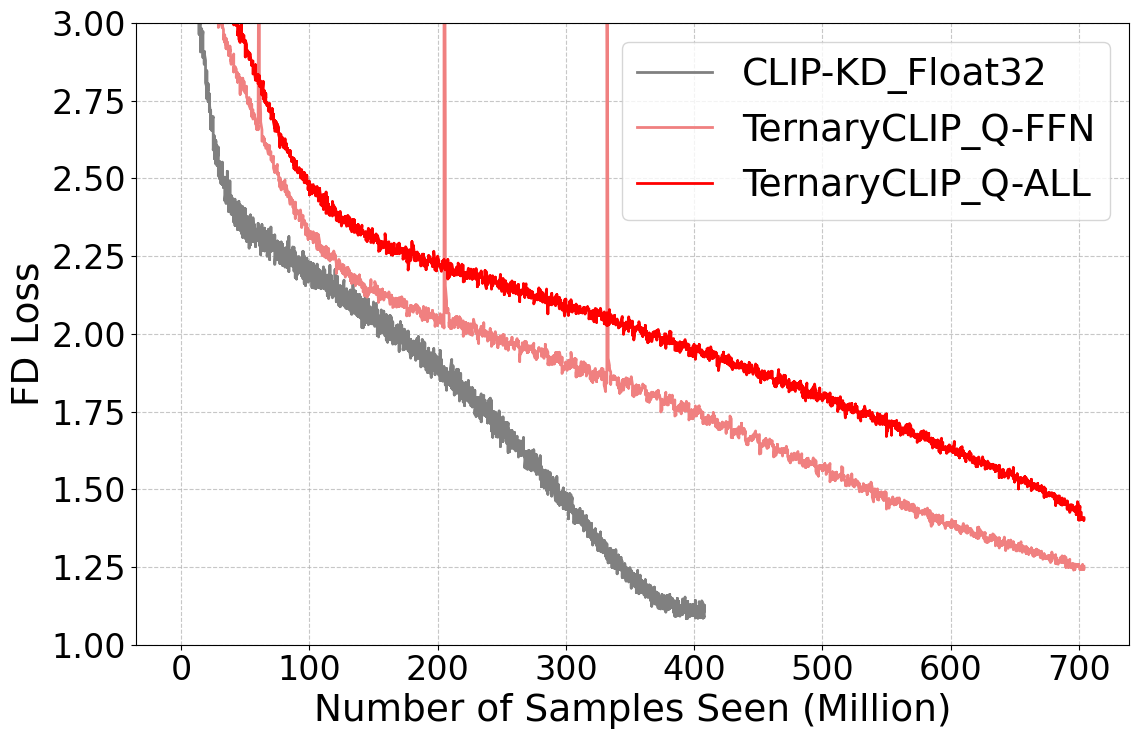}
        \caption{FD losses of full-precision CLIP-KD,  TernaryCLIP\_Q-FFN and TernaryCLIP\_Q-ALL.}
        \label{fig:fd-losses}
    \end{minipage}
\end{figure}

In Figure~\ref{fig:task-losses}, \ref{fig:crd-losses}, \ref{fig:icl-losses}, and \ref{fig:fd-losses}, we show the distinct loss curves of three models on the number of samples seen, as $\mathcal{L}_{\text{total}} = \mathcal{L}_{\text{task}} + \mathcal{L}_{\text{crd}} + \mathcal{L}_{\text{icl}} + \mathcal{L}_{\text{df}}$. The first model is full-precision CLIP-KD, the second model is TernaryCLIP\_Q-FFN with ternarized FFN, and the third model is TernaryCLIP\_Q-ALL with ternarized MHA+FFN. The first loss curve is task loss. The second loss curve is contrastive relational distillation (CRD) loss. The third loss curve is interactive contrastive learning (ICL) loss. The last loss curve is feature distillation (FD) loss. The loss curves are plotted against the number of samples seen during training.

\clearpage

% Details of Zero-Shot Classification -------------------------
\section{Details of Zero-Shot Image Classification}
\label{sec:classification}
% Appendix D. 表格展示所有的 image classification 的 evalution metrics

\begin{figure}[!ht]
    \centering
    \includegraphics[width=\textwidth]{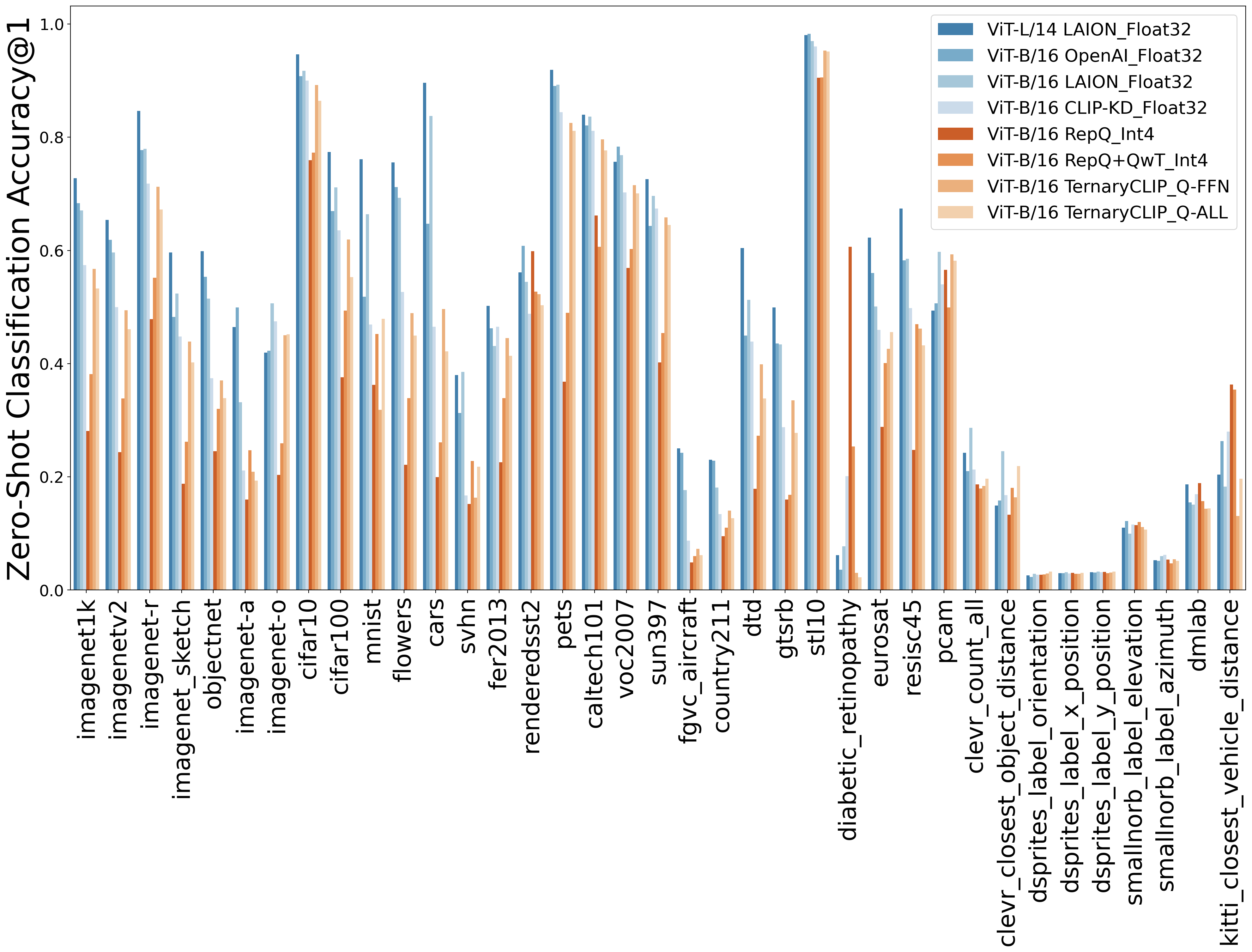}
    \caption{Zero-shot image classification performance: Accuracy@1 across 37 datasets.}
    \label{fig:zero-shot-classification-acc1}
\end{figure}

In Table~\ref{tab:zero_shot_accuracy_details}, TnCLIP represents the abbreviation of TernaryCLIP.
In Figure~\ref{fig:zero-shot-classification-acc1} and Table~\ref{tab:zero_shot_accuracy_details}, TernaryCLIP\_Q-ALL with 99\% ternary weights obtains a 2.71\% performance reduction compared with 32-bit CLIP-KD, and TernaryCLIP\_Q-FFN with 72\% ternary weights gets a 1.83\% performance degradation. Notably, TernaryCLIP\_Q-ALL achieves a 5.29\% performance improvement compared with RepQ+QwT of 4-bit quantized weights and 32-bit compensation weights, and TernaryCLIP\_Q-FFN gains a 6.17\% performance improvement.

\clearpage

\begin{table}[!htb]
\centering
\begin{adjustbox}{max width=\textwidth}
\renewcommand{\arraystretch}{0.7}
\begin{tabular}{l|*{8}{>{\raggedleft\arraybackslash}p{1.8cm}}}
\toprule
\textbf{Datasets} & \textbf{ViT-L/14} & \textbf{ViT-B/16} & \textbf{ViT-B/16} & \textbf{ViT-B/16} & \textbf{ViT-B/16} & \textbf{ViT-B/16} & \textbf{ViT-B/16} & \textbf{ViT-B/16} \\
 & \textbf{LAION} & \textbf{OpenAI} & \textbf{LAION} & \textbf{CLIP-KD} & \textbf{RepQ} & \textbf{RepQ+QwT} & \textbf{TnCLIP} & \textbf{TnCLIP} \\
 & \textbf{\_Float32} & \textbf{\_Float32} & \textbf{\_Float32} & \textbf{\_Float32} & \textbf{\_Int4} & \textbf{\_Int4} & \textbf{\_Q-FFN} & \textbf{\_Q-ALL} \\
\midrule
\multicolumn{9}{c}{\textit{Natural Datasets}} \\
\midrule
caltech101 & 83.99\% & 82.10\% & 83.63\% & 81.15\% & 66.16\% & 60.64\% & 79.61\% & 77.70\% \\
cars & 89.64\% & 64.73\% & 83.77\% & 46.50\% & 19.95\% & 26.07\% & 49.67\% & 42.20\% \\
cifar10 & 94.63\% & 90.77\% & 91.73\% & 89.99\% & 75.96\% & 77.26\% & 89.24\% & 86.45\% \\
cifar100 & 77.39\% & 66.94\% & 71.15\% & 63.57\% & 37.62\% & 49.40\% & 61.96\% & 55.28\% \\
country211 & 23.04\% & 22.87\% & 18.12\% & 13.41\% & 9.52\% & 11.03\% & 14.04\% & 12.68\% \\
dtd & 60.43\% & 44.95\% & 51.28\% & 43.88\% & 17.87\% & 27.29\% & 39.89\% & 33.83\% \\
fer2013 & 50.22\% & 46.22\% & 43.13\% & 46.53\% & 22.58\% & 33.91\% & 44.52\% & 41.40\% \\
fgvc\_aircraft & 25.02\% & 24.24\% & 17.64\% & 8.73\% & 4.89\% & 5.97\% & 7.29\% & 6.15\% \\
flowers & 75.56\% & 71.18\% & 69.28\% & 52.66\% & 22.13\% & 33.92\% & 48.92\% & 44.97\% \\
gtsrb & 49.92\% & 43.56\% & 43.42\% & 28.79\% & 15.98\% & 16.85\% & 33.49\% & 27.75\% \\
imagenet-a & 46.49\% & 49.92\% & 33.19\% & 21.12\% & 15.97\% & 24.69\% & 20.91\% & 19.36\% \\
imagenet-o & 41.95\% & 42.30\% & 50.65\% & 47.45\% & 20.35\% & 25.95\% & 45.00\% & 45.20\% \\
imagenet-r & 84.68\% & 77.71\% & 77.93\% & 71.83\% & 47.84\% & 55.20\% & 71.28\% & 67.22\% \\
imagenet1k & 72.77\% & 68.36\% & 67.07\% & 57.44\% & 28.10\% & 38.17\% & 56.76\% & 53.28\% \\
imagenetv2 & 65.41\% & 61.90\% & 59.65\% & 50.01\% & 24.38\% & 33.85\% & 49.43\% & 46.10\% \\
objectnet & 59.86\% & 55.33\% & 51.49\% & 37.41\% & 24.56\% & 32.00\% & 37.02\% & 33.89\% \\
pets & 91.91\% & 89.04\% & 89.26\% & 84.44\% & 36.82\% & 49.01\% & 82.56\% & 81.14\% \\
stl10 & 98.05\% & 98.25\% & 96.99\% & 96.05\% & 90.53\% & 90.56\% & 95.29\% & 95.16\% \\
sun397 & 72.59\% & 64.34\% & 69.61\% & 67.41\% & 40.23\% & 45.43\% & 65.84\% & 64.48\% \\
svhn & 37.97\% & 31.27\% & 38.53\% & 16.72\% & 15.19\% & 22.80\% & 16.33\% & 21.79\% \\
voc2007 & 75.64\% & 78.32\% & 76.85\% & 70.23\% & 56.92\% & 60.26\% & 71.51\% & 70.10\% \\
\midrule
\multicolumn{9}{c}{\textit{Specialized Datasets}} \\
\midrule
diabetic\_retinopathy & 6.19\% & 3.57\% & 7.75\% & 20.09\% & 60.67\% & 25.36\% & 3.03\% & 2.27\% \\
eurosat & 62.24\% & 56.00\% & 50.07\% & 45.96\% & 28.85\% & 40.09\% & 42.59\% & 45.59\% \\
imagenet\_sketch & 59.65\% & 48.24\% & 52.37\% & 44.79\% & 18.81\% & 26.22\% & 43.91\% & 40.23\% \\
mnist & 76.10\% & 51.85\% & 66.39\% & 46.90\% & 36.27\% & 45.25\% & 31.86\% & 47.93\% \\
pcam & 49.38\% & 50.67\% & 59.73\% & 53.98\% & 56.58\% & 49.94\% & 59.30\% & 58.18\% \\
renderedsst2 & 56.12\% & 60.79\% & 54.48\% & 48.82\% & 59.86\% & 52.72\% & 52.28\% & 50.30\% \\
resisc45 & 67.43\% & 58.27\% & 58.54\% & 49.83\% & 24.73\% & 46.95\% & 46.21\% & 43.24\% \\
\midrule
\multicolumn{9}{c}{\textit{Structured Datasets}} \\
\midrule
clevr\_closest\_object\_distance & 14.91\% & 15.83\% & 24.51\% & 16.75\% & 13.29\% & 18.07\% & 16.40\% & 21.90\% \\
clevr\_count\_all & 24.25\% & 21.03\% & 28.65\% & 21.27\% & 18.65\% & 17.93\% & 18.37\% & 19.65\% \\
dmlab & 18.66\% & 15.50\% & 15.10\% & 16.92\% & 18.88\% & 15.70\% & 14.39\% & 14.40\% \\
dsprites\_label\_orientation & 2.61\% & 2.34\% & 2.87\% & 2.71\% & 2.71\% & 2.77\% & 2.92\% & 3.28\% \\
dsprites\_label\_x\_position & 2.99\% & 3.00\% & 3.15\% & 2.93\% & 3.05\% & 2.90\% & 2.86\% & 3.04\% \\
dsprites\_label\_y\_position & 3.16\% & 3.11\% & 3.24\% & 3.16\% & 3.20\% & 3.01\% & 3.07\% & 3.28\% \\
kitti\_closest\_vehicle\_distance & 20.39\% & 26.30\% & 18.28\% & 27.99\% & 36.29\% & 35.44\% & 13.08\% & 19.69\% \\
smallnorb\_label\_azimuth & 5.25\% & 5.18\% & 6.02\% & 6.20\% & 5.39\% & 4.71\% & 5.46\% & 5.16\% \\
smallnorb\_label\_elevation & 11.00\% & 12.21\% & 9.96\% & 11.59\% & 11.49\% & 12.00\% & 11.11\% & 10.71\% \\
\midrule
\multicolumn{9}{c}{\textit{Summary}} \\
\midrule
Average (natural) & 65.58\% & 60.68\% & 61.16\% & 52.16\% & 33.03\% & 39.06\% & 51.46\% & 48.86\% \\
Average (specialized) & 53.87\% & 47.06\% & 49.90\% & 44.34\% & 40.82\% & 40.93\% & 39.88\% & 41.11\% \\
Average (structured) & 11.47\% & 11.61\% & 12.42\% & 12.17\% & 12.55\% & 12.50\% & 9.74\% & 11.23\% \\
Average (All) & 50.20\% & 46.17\% & 47.18\% & 40.95\% & 29.52\% & 32.95\% & 39.12\% & 38.24\% \\
Perf vs. ViT-L/14 LAION & 0.00\% & -4.03\% & -3.02\% & -9.25\% & -20.68\% & -17.25\% & -11.08\% & -11.96\% \\
Perf vs. ViT-B/16 OpenAI & 4.03\% & 0.00\% & 1.01\% & -5.22\% & -16.65\% & -13.22\% & -7.05\% & -7.93\% \\
Perf vs. ViT-B/16 LAION & 3.02\% & -1.01\% & 0.00\% & -6.23\% & -17.66\% & -14.23\% & -8.06\% & -8.94\% \\
Perf vs. ViT-B/16 CLIP-KD & 9.25\% & 5.22\% & 6.23\% & 0.00\% & -11.43\% & -8.00\% & \textbf{-1.83\%} & \textbf{-2.71\%} \\
Perf vs. ViT-B/16 RepQ & 20.68\% & 16.65\% & 17.66\% & 11.43\% & 0.00\% & 3.43\% & \textbf{9.60\%} & \textbf{8.72\%} \\
Perf vs. ViT-B/16 RepQ+QwT & 17.25\% & 13.22\% & 14.23\% & 8.00\% & -3.43\% & 0.00\% & \textbf{6.17\%} & \textbf{5.29\%} \\
\bottomrule
\end{tabular}
\end{adjustbox}
\caption{Zero-shot image classification performance: Accuracy@1 on 37 datasets covering 3 distinct types.}
\label{tab:zero_shot_accuracy_details}
\end{table}

\clearpage

% Details of Inference Latency ------------------------------------------
\section{Details of Inference Latency}
\label{sec:latency}
% Appendix E. bits per weight: bpw, 各种环境下的细致的时延对比

In Table~\ref{tab:latency-m4-pro}, we present a comprehensive analysis of CLIP model inference latency across various precision formats with the corresponding practical bit-per-weight (BPW) implementations by the GGML library, indicating the decomposition of total latency into model loading, image loading, and model forwarding components, with all benchmarks conducted on Apple M4 Pro hardware over 1,000 test rounds.

% TAB: Latency
\begin{table}[!htb]
\centering
\begin{minipage}{\textwidth}
\begin{adjustbox}{max width=\textwidth}
\begin{tabular}{l|lcccccc}
\hline
\textbf{Models} & \textbf{Precision} & \textbf{BPW}~$\downarrow$ & \textbf{Storage(MB)}~$\downarrow$ & \textbf{Model Load(ms)}~$\downarrow$ & \textbf{Image Load(ms)}~$\downarrow$ & \textbf{Model Forward(ms)}~$\downarrow$ & \textbf{Total Latency(ms)}~$\downarrow$ \\
\hline
\multirow{2}{*}{ViT-L/14} & Float32 & 32 & 1630.90 & 382.22 $\pm$ 22.58 & 5.32 $\pm$ 1.65 & 498.60 $\pm$ 85.66 & 886.15 $\pm$ 90.65 \\
 & Float16 & 16 & 817.28 & 199.42 $\pm$ 9.50 & 5.35 $\pm$ 0.24 & 506.10 $\pm$ 75.54 & 710.88 $\pm$ 76.38 \\
\hline
\multirow{2}{*}{ViT-B/16} & Float32 & 32 & 571.60 & 153.02 $\pm$ 7.41 & 5.35 $\pm$ 0.22 & 83.02 $\pm$ 55.47 & 241.40 $\pm$ 56.47 \\
 & Float16 & 16 & 287.73 & 90.15 $\pm$ 4.27 & 5.42 $\pm$ 0.63 & 79.04 $\pm$ 50.26 & 174.62 $\pm$ 50.52 \\
\hline
\multirow{6}{*}{TernaryCLIP\_Q\_FFN ViT-B/16} & Float32 & 32 & 571.60 & 154.55 $\pm$ 21.00 & 5.79 $\pm$ 0.77 & 90.03 $\pm$ 59.39 & 250.38 $\pm$ 67.65 \\
 & Float16 & 16 & 287.73 & 90.52 $\pm$ 4.87 & 5.84 $\pm$ 0.72 & 86.62 $\pm$ 53.00 & 182.99 $\pm$ 53.57 \\
 & Q4\_0 & 4 & 140.93 & 57.32 $\pm$ 4.35 & 5.83 $\pm$ 0.44 & 74.02 $\pm$ 49.97 & 137.18 $\pm$ 50.52 \\
 & Q4\_1 & 4 & 147.31 & 58.53 $\pm$ 2.86 & 5.82 $\pm$ 0.38 & 70.40 $\pm$ 48.54 & 134.75 $\pm$ 48.68 \\
 & TQ1\_0 & 1.6875 & 105 & 49.09 $\pm$ 3.47 & 5.81 $\pm$ 0.33 & 71.14 $\pm$ 50.89 & 126.05 $\pm$ 51.05 \\
%  & TQ2\_0 & 2.0625 & 109.8 & 50.15 $\pm$ 3.00 & 5.82 $\pm$ 0.38 & 76.26 $\pm$ 50.06 & 132.23 $\pm$ 49.99 \\
\hline
\multirow{6}{*}{TernaryCLIP\_Q\_ALL ViT-B/16} & Float32 & 32 & 571.60 & 149.82 $\pm$ 11.73 & 5.63 $\pm$ 0.45 & 87.83 $\pm$ 56.36 & 243.28 $\pm$ 57.14 \\
 & Float16 & 16 & 287.73 & 89.12 $\pm$ 3.77 & 5.77 $\pm$ 0.32 & 79.70 $\pm$ 50.35 & 174.60 $\pm$ 50.40 \\
 & Q4\_0 & 4 & 84.87 & 44.16 $\pm$ 1.79 & 5.75 $\pm$ 0.24 & 71.30 $\pm$ 49.68 & 121.22 $\pm$ 49.79 \\
 & Q4\_1 & 4 & 93.69 & 46.10 $\pm$ 1.87 & 5.75 $\pm$ 0.29 & 68.46 $\pm$ 46.07 & 120.32 $\pm$ 46.08 \\
 & TQ1\_0 & \textbf{1.6875} & \textbf{35.25} & \textbf{33.05 $\pm$ 1.70} & \textbf{5.74 $\pm$ 0.19} & \textbf{67.96 $\pm$ 45.58} & \textbf{106.75 $\pm$ 45.68} \\
%  & TQ2\_0 & 2.0625 & 41.9 & 34.77 $\pm$ 2.75 & 5.78 $\pm$ 0.45 & 75.78 $\pm$ 48.16 & 116.33 $\pm$ 48.21 \\
\hline
\end{tabular}
\end{adjustbox}
\caption{CLIP model inference latency overview on different precisions and bpw (bits per weight). The CPU hardware is Apple M4 Pro. Total latency is combined with model loading, image loading, and model forwarding. Latency = average value $\pm$ three times standard deviation under 1,000 rounds of benchmarks.}
\label{tab:latency-m4-pro}
\end{minipage}
\end{table}

\clearpage

% Ablation Study --------------------------------------------------------
\section{Ablation Study}
\label{sec:ablation-study}
% Appendix F. 为什么不用Data augment % 为什么没有激活值量化 % 是否使用蒸馏的方式有什么区别 % self-distill 的差别也不大

\begin{figure}[!ht]
    \centering
    \begin{minipage}[t]{0.48\textwidth}
        \centering
        \includegraphics[width=\linewidth]{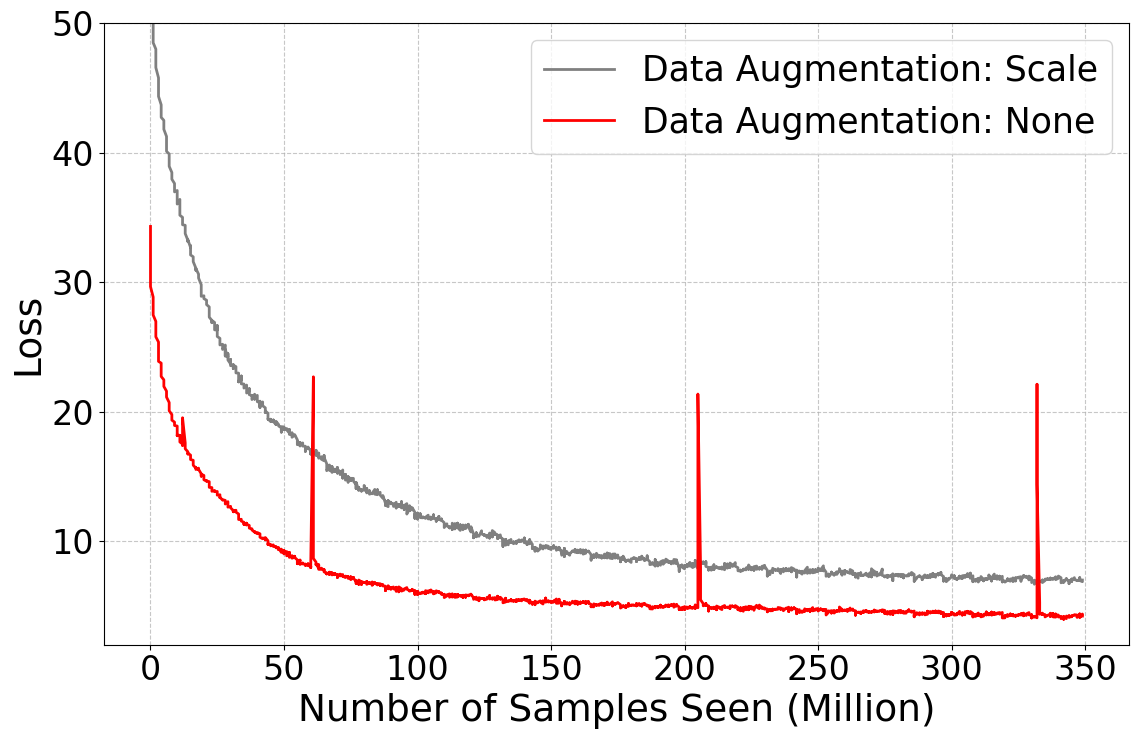}
        \caption{Ablation study of data augmentation on TernaryCLIP. Data augmentation makes loss convergence much slower than the baseline.}
        \label{fig:ablation-data-aug}
    \end{minipage}
    \hfill
    \begin{minipage}[t]{0.48\textwidth}
        \centering
        \includegraphics[width=\linewidth]{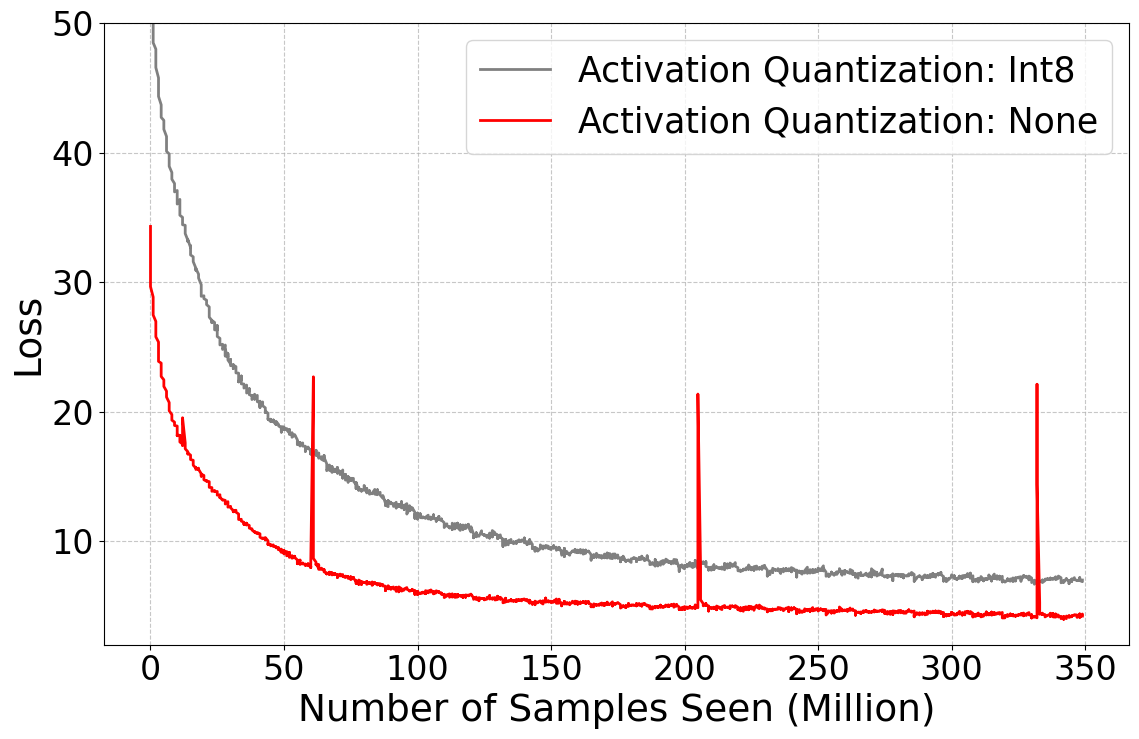}
        \caption{Ablation study of int8 or float16 activation quantization on TernaryCLIP. The int8 activation quantization makes loss convergence much slower than the baseline.}
        \label{fig:ablation-act-quant}
    \end{minipage}
\end{figure}

In Figure~\ref{fig:ablation-data-aug} and Figure~\ref{fig:ablation-act-quant}, we conducted an ablation study of data augmentation and int8 activation quantization. Based on the performance gap of loss curves, we determine training TernaryCLIP without any data augmentation and activation quantization.

\begin{figure}[!ht]
    \centering
    \begin{minipage}[t]{0.48\textwidth}
        \centering
        \includegraphics[width=\linewidth]{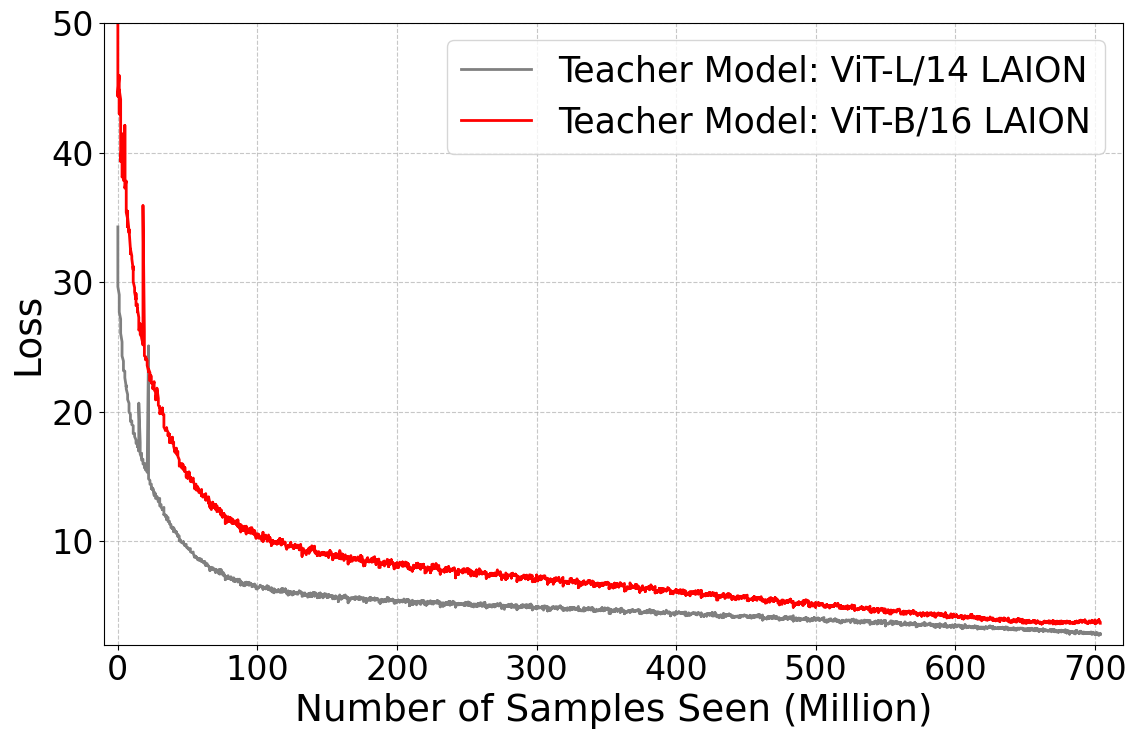}
        \caption{Ablation study of self-distillation and distillation on TernaryCLIP training. Self-distillation maintains similar convergence characteristics.}
        \label{fig:ablation-self-distill-loss}
    \end{minipage}
    \hfill
    \begin{minipage}[t]{0.48\textwidth}
        \centering
        \includegraphics[width=\linewidth]{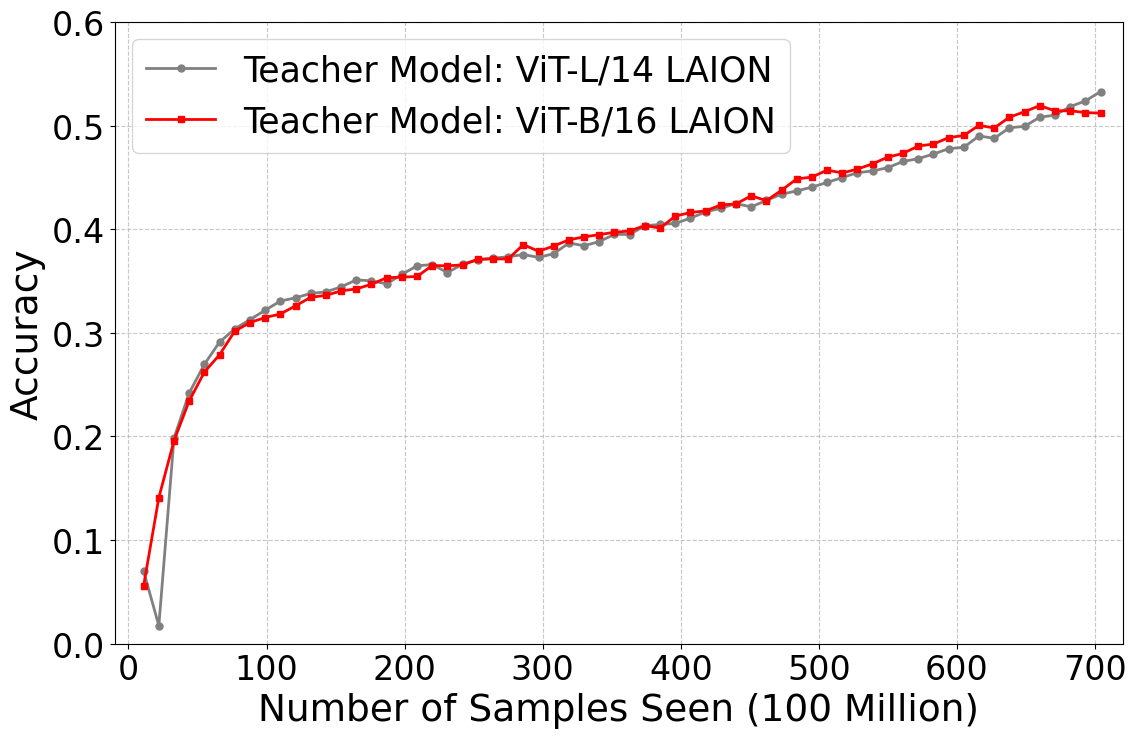}
        \caption{Ablation study of self-distillation and distillation on TernaryCLIP accuracy. Self-distillation achieves only 1.35\% performance degradation.}
        \label{fig:ablation-self-distill-acc}
    \end{minipage}
\end{figure}

In Figure~\ref{fig:ablation-self-distill-loss} and Figure~\ref{fig:ablation-self-distill-acc}, we conducted an ablation study on the effect of self-distillation and distillation with different teacher models. The performance degradation of self-distillation is only 1.35\% compared with distillation, which is acceptable for the sake of training efficiency and simplicity. The results show that self-distillation achieves competitive performance as distillation under the same training configurations, which indicates that ternarization-aware distillation is not limited to the larger teacher model.
% 增加一个 self-distill 和 distill 的对比

\clearpage

\section{Ternary Weight Distribution}
\label{sec:ternary-weight-distribution}
% Appendix G. 画出三元量化权重分布的图像

\begin{figure*}[!ht]
\begin{minipage}{\textwidth}
\vspace{0.3cm}
\centering
\includegraphics[width=\textwidth]{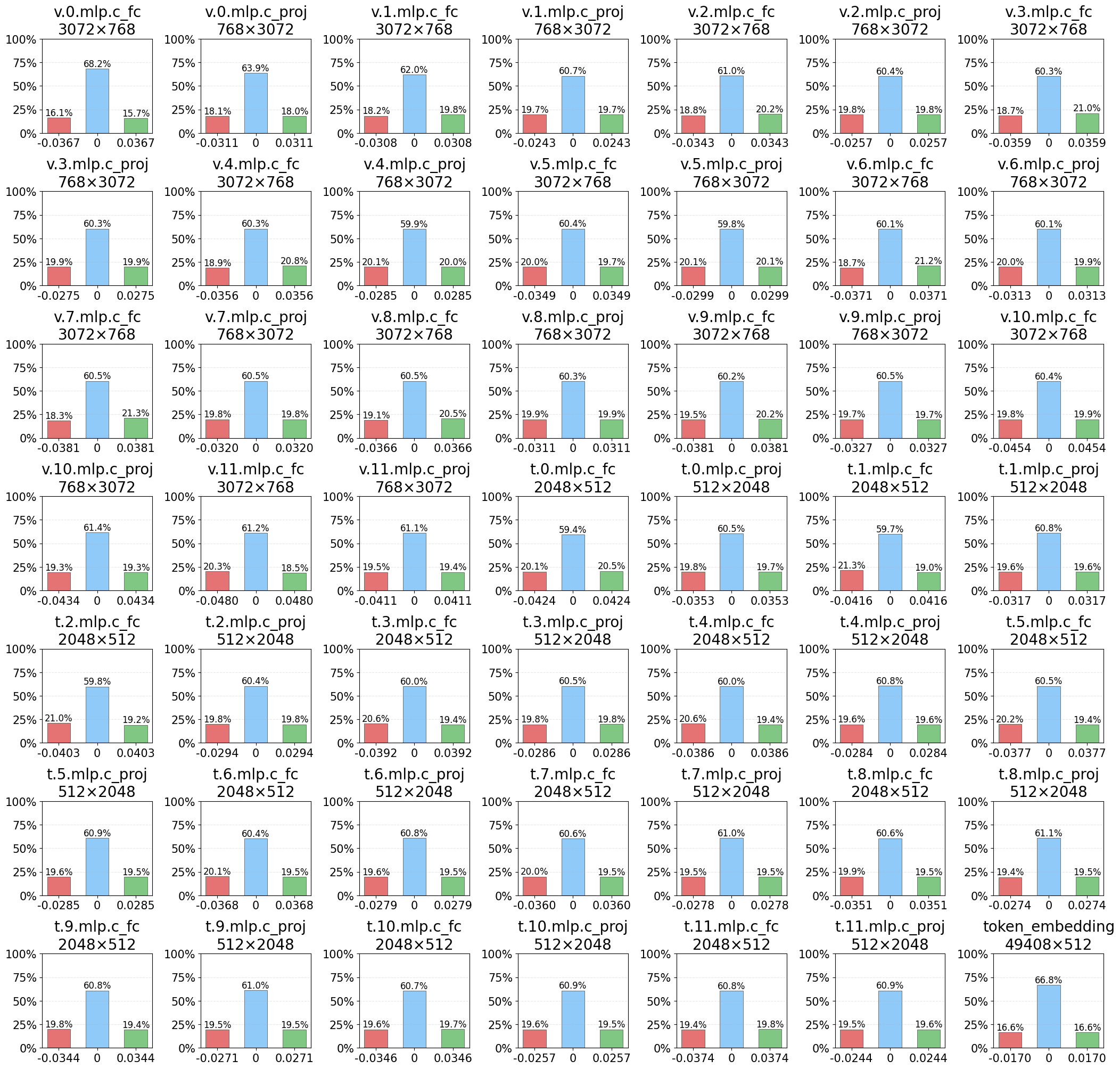}
\caption{Ternary weight distribution of TernaryCLIP\_Q-FFN.}
\label{fig:w-dist-qffn}
\end{minipage}
\end{figure*}

\begin{figure*}[!t]
\begin{minipage}{\textwidth}
\vspace{0.3cm}
\centering
\includegraphics[width=\textwidth]{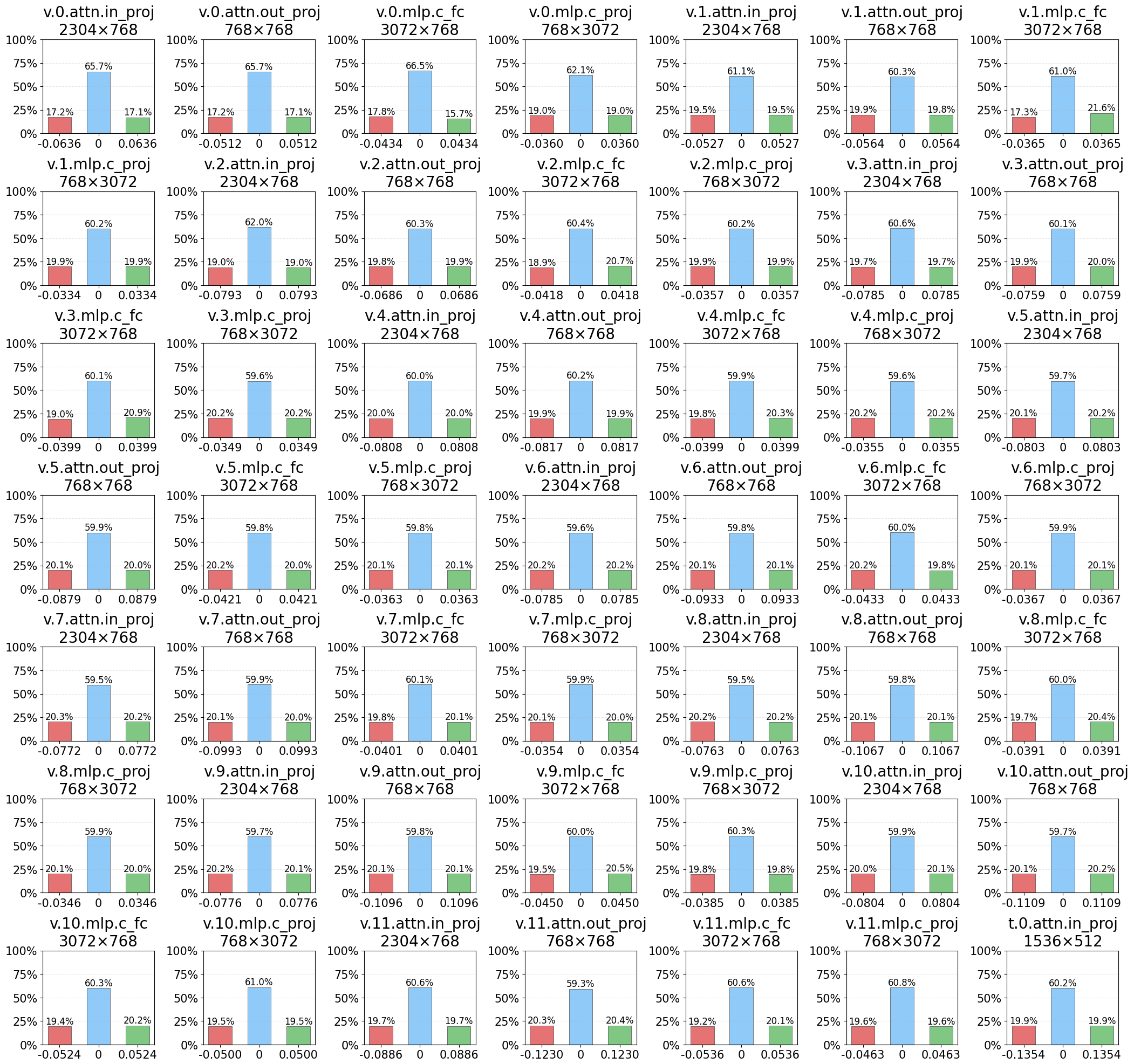}
\caption{Ternary weight distribution of TernaryCLIP\_Q-ALL (Part1).}
\label{fig:w-dist-qall-part0}
\end{minipage}
\end{figure*}

\begin{figure*}[!t]
\begin{minipage}{\textwidth}
\vspace{0.3cm}
\centering
\includegraphics[width=\textwidth]{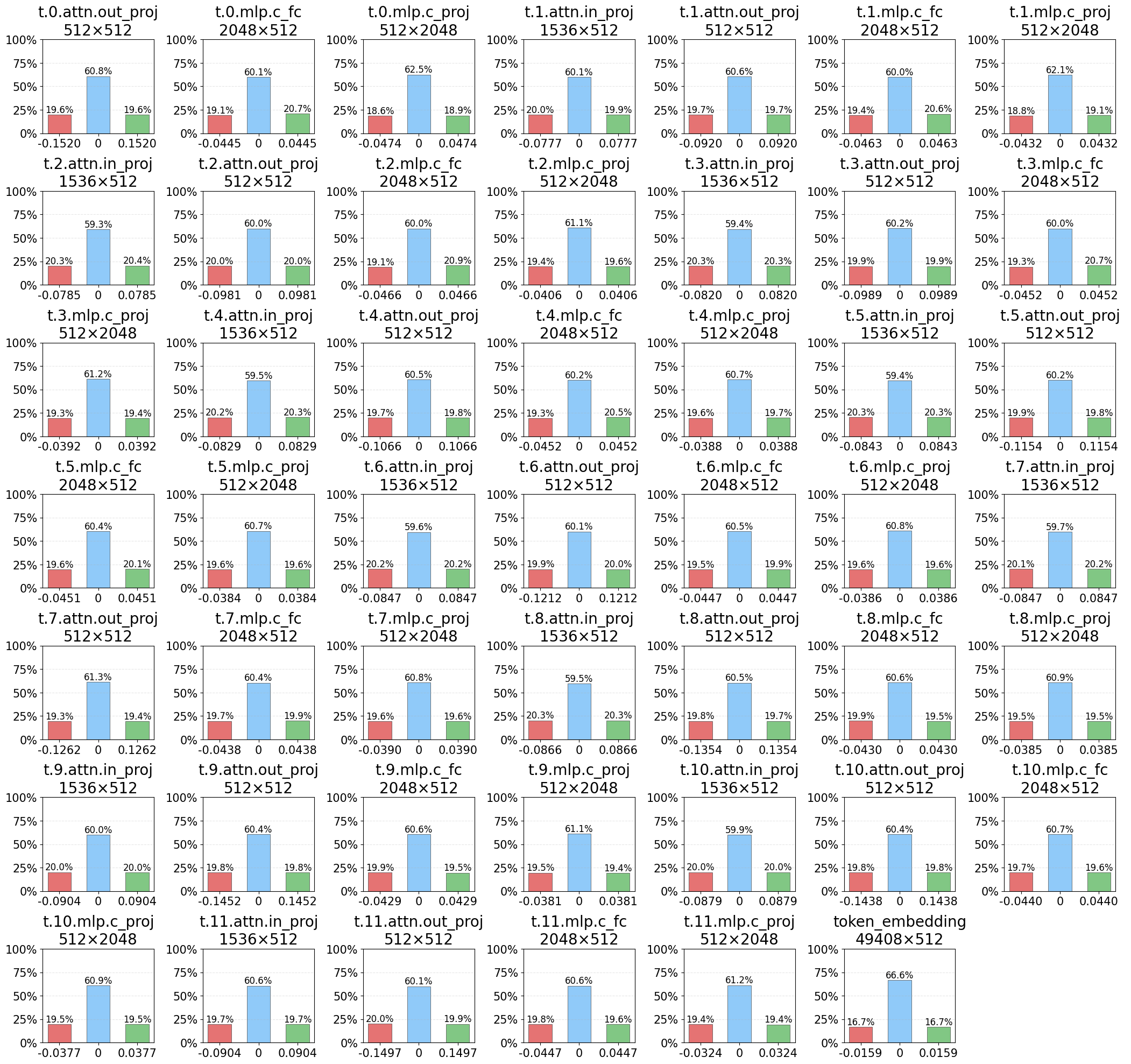}
\caption{Ternary weight distribution of TernaryCLIP\_Q-ALL (Part2).}
\label{fig:w-dist-qall-part1}
\end{minipage}
\end{figure*}

In Figure~\ref{fig:w-dist-qffn},~Figure~\ref{fig:w-dist-qall-part0},~and~Figure~\ref{fig:w-dist-qall-part1}, we illustrate the distribution of ternary weight for TernaryCLIP Q-FFN and Q-ALL models. It is obvious that ternary weights maintain a good sparsity in addition to other benefits of quantization that we have discussed during experiments.

% --------------------------------------------------------------------
\end{document}